\pdfoutput=1

\documentclass[11pt]{article}

\usepackage[final]{acl}

\usepackage{times}
\usepackage{latexsym}
\usepackage{subcaption}
\usepackage{placeins}
\usepackage[T1]{fontenc}

\usepackage[utf8]{inputenc}
\usepackage{lipsum}
\usepackage{newunicodechar}
\usepackage{hhline}
\usepackage{multirow}
\usepackage{tabularx}
\usepackage{longtable}
\newcolumntype{Y}{>{\centering\arraybackslash}X}
\newcolumntype{Z}{>{\raggedleft\arraybackslash}X}
\usepackage{hyperref}

\newunicodechar{⁠}{} 
\usepackage{microtype}
\usepackage{makecell}
\usepackage{inconsolata}

\usepackage{graphicx}

\title{\texttt{SiDiaC}: Sinhala Diachronic Corpus}

\author{Nevidu Jayatilleke \and Nisansa de Silva \\
  Department of Computer Science \& Engineering, 
  University of Moratuwa, 
  Sri Lanka \\
  \texttt{\{nevidu.25, NisansaDdS\}@cse.mrt.ac.lk} 
  }

\begin{document}
\maketitle
\begin{abstract}
\texttt{SiDiaC}, the first comprehensive \textit{Sinhala Diachronic Corpus}, covers a historical span from the 5th to the 20th century CE. \texttt{SiDiaC} comprises 58k words across 46 literary works, annotated carefully based on the written date, after filtering based on availability, authorship, copyright compliance, and data attribution. Texts from the \textit{National Library of Sri Lanka} were digitised using \texttt{Google Document AI} OCR engine, followed by post-processing to correct formatting and modernise the orthography. The construction of \texttt{SiDiaC} was informed by practices from other corpora, such as \textit{FarPaHC}, particularly in syntactic annotation and text normalisation strategies, due to the shared characteristics of low-resourced language status. 
This corpus is categorised based on genres into two layers: primary and secondary. Primary categorisation is binary, classifying each book into Non-Fiction or Fiction, while the secondary categorisation is more specific, grouping texts under Religious, History, Poetry, Language, and Medical genres. 
%
%
Despite challenges including limited access to rare texts and reliance on secondary date sources, \texttt{SiDiaC} serves as a foundational resource for Sinhala NLP, significantly extending the resources available for Sinhala, enabling diachronic studies in lexical change, neologism tracking, historical syntax, and corpus-based lexicography. 
\end{abstract}

\section{Introduction}

\textit{Sola lingua bona est lingua mortua}\footnote{\textbf{Latin:} The only good language is a dead language.}; given that all languages that are in use, evolve with a gradual process of linguistic change over time, proposed to have originated from an initially gestural communication system~\cite{corballis2017evolution}.
Factors affecting this complex evolution include cultural influences, which drive irregular word meaning shifts through phenomena such as new technologies (e.g., \textit{cell} to mean \textit{cell phone} in addition to \textit{prison cell}) or community-specific vernaculars (e.g., \textit{gay} to mean \textit{homosexual} in addition to \textit{carefree}). These cultural shifts often impact nouns more significantly. Conversely, regular linguistic processes, such as subjectification (e.g., \textit{actually} shifting from objective to subjective usage) or grammaticalisation (e.g., \textit{promise} undergoing rich changes), cause more predictable semantic changes and tend to affect verbs, adjectives, and adverbs more readily~\cite{hamilton-etal-2016-cultural}. 

The Sinhala language is an Indo-European language, which possesses a rich and diverse literary heritage that has developed over the course of several millennia, with its origins tracing back to between the 3rd and 2nd centuries BCE ~\cite{de2025survey}. This language has undergone significant evolution and transformation throughout its history, resulting in the form of modern Sinhala that we engage with today. 
%
%
Sinhala is spoken as L1 by approximately 16 million people, primarily located on the island of Sri Lanka~\cite{de2025survey}. The Sinhala script, which is unique to the language, descends from the Indian Brahmi script~\cite{fernando1949palaeographical,de-mel-etal-2025-sinhala}.
Sinhala is classified as a lower-resourced language (Category 02) according to the criteria presented by~\citet{ranathunga-de-silva-2022-languages}.

In this study, we introduce a novel diachronic Sinhala dataset, \texttt{SiDiaC}\footnote{\label{note: sidiac} \scriptsize \urlstyle{tt}\url{https://github.com/NeviduJ/SiDiaC}}, which covers the period from 426 CE to 1944 CE. This dataset is based on distinct identifications of written years or specific time frames of the recognised Sinhala literature.

\section{Existing Work}

The development of historical corpora has garnered significant attention due to its importance beyond the creation of general-purpose corpora. It enables researchers to investigate the evolution of language, taking into account changes in semantics, lexicon, morphology, and syntax. As a result, studies have been conducted to develop diachronic corpora for different languages.

\subsection{\texttt{LatinlSE}}

\citet{mcgillivray2013tools} introduced \texttt{LatinlSE}, a 13-million-word historical Latin corpus developed for the Sketch Engine\footnote{\label{note: sketcheng} \scriptsize \urlstyle{tt}\url{https://www.sketchengine.eu/}}, a leading corpus query tool. 
Covering an extensive 22-century period from the 2nd Century BCE to the 21st Century CE, \texttt{LatinlSE} is equipped with detailed metadata, including author, title, genre, era, date, and century. 

The methodology for creating this corpus involved gathering texts from various online digital libraries, such as \textit{LacusCurtius}\footnote{\scriptsize \urlstyle{tt}\url{https://penelope.uchicago.edu/Thayer/E/Roman/Texts/}}, \textit{IntraText}\footnote{\label{note: intratext} \scriptsize \urlstyle{tt}\url{https://www.intratext.com/}}, and \textit{Musisque Deoque}\footnote{\label{note: mqdq} \scriptsize \urlstyle{tt}\url{https://www.mqdq.it/}}. This process ensured a broad classification of genres as prose and poetry, and the texts were converted into a verticalised format while preserving their metadata. A significant aspect of the creation process was the automatic linguistic annotation using advanced NLP tools. This included lemmatisation with the \texttt{PROIEL}\footnote{\label{note: proiel} \scriptsize \urlstyle{tt}\url{https://www.hf.uio.no/ifikk/english/research/projects/proiel/}} project's morphological analyser, complemented by \textit{Quick Latin}\footnote{\label{note: quick_lat} \scriptsize \urlstyle{tt}\url{https://www.quicklatin.com/}} for unrecognised forms. Part-of-Speech (POS) tagging was achieved by training \texttt{TreeTagger}~\cite{schmid1999improvements} on existing Latin treebanks, including the \textit{Index Thomisticus Treebank}\footnote{\label{note: treebank} \scriptsize \urlstyle{tt}\url{https://itreebank.marginalia.it/}}, the \textit{Latin Dependency Treebank}~\cite{bamman2006design}, and the \texttt{PROIEL} project's Latin treebank~\cite{haug2008creating}. This training helps disambiguate analyses and assign the most likely lemma and POS to each token in context. This comprehensive dataset allows users to perform sophisticated searches based on lemmas, POS, and context, facilitating the study of shifts in word meanings over time.

\subsection{\texttt{IcePaHC} and \texttt{FarPaHC}}

The \textit{Icelandic Parsed Historical Corpus} (\texttt{IcePaHC})~\cite{rognvaldsson-etal-2012-icelandic} is a one-million-word parsed historical corpus of Icelandic, spanning from the late 12th century to the early 21st century. 
But more relevant to our work in this study is the \textit{Faroese Parsed Historical Corpus} (\texttt{FarPaHC}), a syntactically annotated corpus of Faroese historical texts, that is presented as a \textit{spin-off} of \texttt{IcePaHC}. The reason for this relevance is that, according to~\citet{ranathunga-de-silva-2022-languages}, Faroese also belongs to Category 02, similar to Sinhala. The \texttt{FarPaHC} corpus has 53,000 words. 

It's given that \texttt{FarPaHC} is an extension of \texttt{IcePaHC}; the primary sources included narrative and religious texts that have parallel texts in \texttt{IcePaHC}.  
A key step in the process was the conversion of all texts to modern spelling using the \texttt{IceNLP package}\footnote{\label{note: icenlp} \scriptsize \urlstyle{tt}\url{https://sourceforge.net/projects/icenlp/}} (which includes a tokeniser, POS tagger, and lemmatiser), which was necessary for preprocessing and for facilitating searches. 
The annotation process involved manually dividing clauses, semi-automatically preprocessing texts with \texttt{IceNLP} and \texttt{CorpusSearch}\footnote{\label{note: corpsearch} \scriptsize \urlstyle{tt}\url{https://corpussearch.sourceforge.net/}} for partial annotations, and extensive manual parsing carried out by one annotator using a custom-developed visual tree editor, \texttt{Annotald}\footnote{\label{note: annotald} \scriptsize \urlstyle{tt}\url{https://github.com/Annotald/annotald}}.

\subsection{Other Historical Corpora}

\citet{pettersson2019characteristics} provides a comprehensive survey on existing diachronic and historical corpora.  
The work by~\citet{keersmaekers2024creating} presents a case study demonstrating how large-scale automated parsing of Greek papyri can create richly annotated diachronic resources. \citet{chen2025administrative} have created a Chinese corpus from the last 30 years of news articles on land usage.
Even the corpora in higher-resourced languages such as \texttt{DIAKORP}~\cite{kuvcera2015diakorp} (Czech), \texttt{ARCHER}~\cite{biber1994archer}, \texttt{COHA}~\cite{davies2012expanding}\footnote{\label{note: coha} \scriptsize \urlstyle{tt}\url{https://www.english-corpora.org/coha/}} (English), \texttt{DTA}~\cite{geyken2011deutsche} and \texttt{GerManC}~\cite{scheible-etal-2011-gold} (German) differ in size, balance, annotation depth, and access models. \texttt{DIAKORP} offers seven centuries of Czech texts, though it lacks linguistic annotation. \texttt{ARCHER} samples English registers across four centuries in 50-year intervals, while \texttt{COHA} spans two centuries of American English with lemmatisation and POS tagging.
\textit{Penn Parsed Corpora of Historical English}~\cite{taylor1994penn} (\texttt{PPCHE}) and \texttt{SRCMF}~\cite{prevost2013syntactic} (Old French) are similar to \texttt{FarPaHC} in the sense that they, too, are manually annotated corpora which provide syntactic analyses suitable for structural studies. \texttt{PPCHE}, in particular, has influenced corpora in other languages through its \textit{Penn-Helsinki} annotation scheme, facilitating cross-linguistic comparison. Similarly, the \texttt{PROIEL} treebank family~\cite{eckhoff2018proiel} extends such comparisons to some Indo-European languages via aligned New Testament translations.

\texttt{ReM}~\cite{klein2016handbuch} (Middle High German), \texttt{RIDGES}~\cite{odebrecht2017ridges} (German-Science), and the \textit{Swedish Culturomics Gigaword} corpus~\cite{eide2016swedish}, offer layered annotation or harmonised spellings for OCR quality control. While all corpora use some metadata scheme to provide critical contextual information such as date, genre, region, and authorship, beyond that, the metadata coverage varies widely. However, it can be noted that TEI-based\footnote{\label{note: coha} \scriptsize \urlstyle{tt}\href{https://tei-c.org/guidelines/}{Text Encoding Initiative (TEI) guidelines}} metadata schemes are popular among European language corpora.



\section{Methodology}

\begin{figure}[!htbp]
    \centering
    \setlength{\fboxsep}{1pt} 
    \setlength{\fboxrule}{0.4pt} 
    \fbox{%
        \includegraphics[width=0.8\columnwidth]{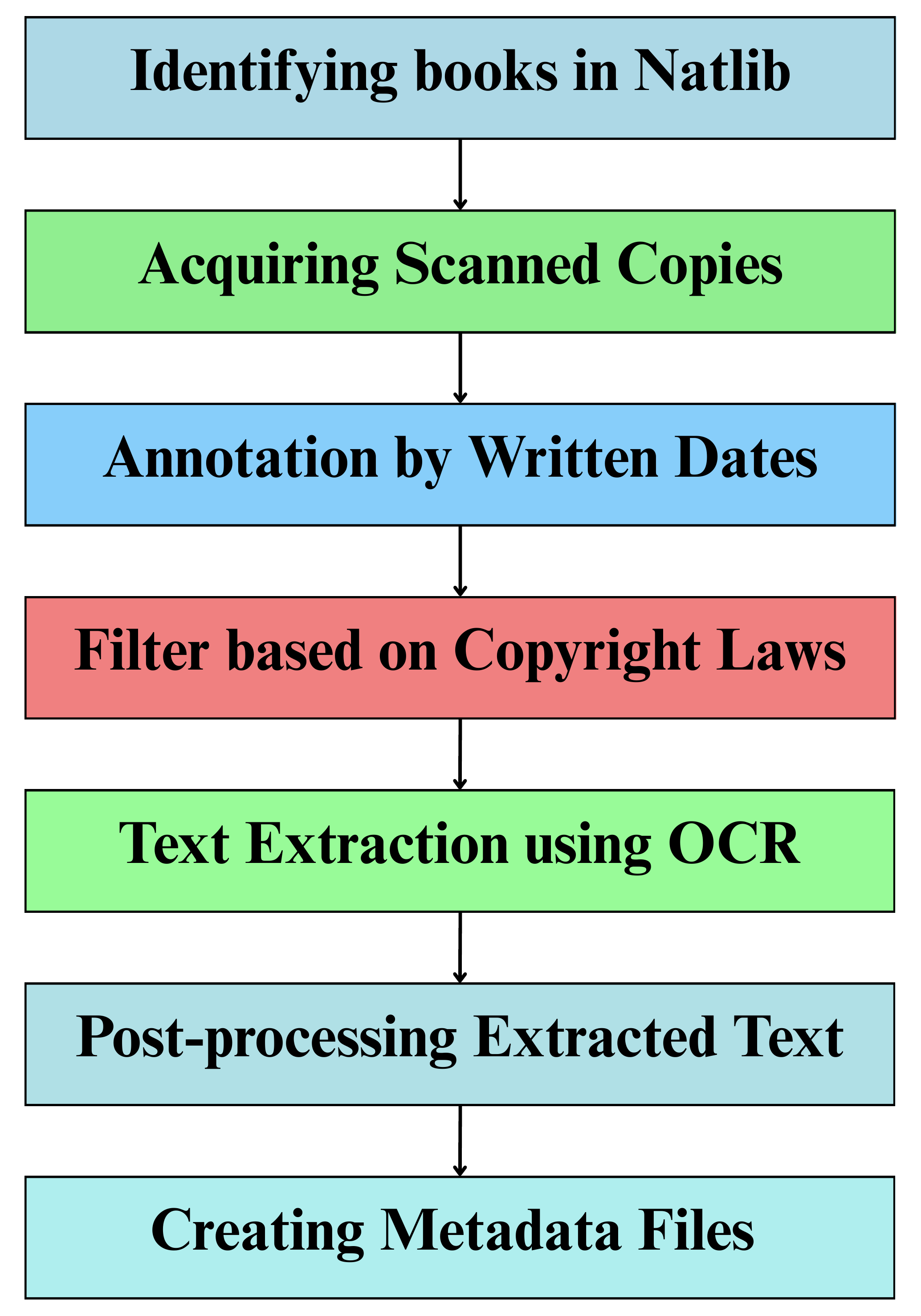}%
    }%
    \caption{Summary of the Methodology Used in the Creation of \texttt{SiDiaC}.}
    \label{fig:summMethod}
\end{figure}

In this section, we describe the methodology used to create this dataset from the ground up. The process involved careful attention to detail at every stage, from planning to the final presentation, ensuring that the data were valid and of high quality. The procedure included addressing copyright laws in Sri Lanka, acquiring data, extracting text, and performing post-processing and formatting of the data as shown in Figure~\ref{fig:summMethod}.

\subsection{Dataset Assembly}

At first, we began acquiring Sinhala literature, including both fiction and non-fiction books, from the \textit{Internet Archives}. However, the amount of data we were able to gather was quite limited. As a result, we decided to turn to the primary institution dedicated to Sinhala literature: the National Library (\texttt{Natlib}) of Sri Lanka\footnote{\label{note: natlib} \scriptsize \urlstyle{tt}\url{https://www.natlib.lk/}}, which has its own digital repository\footnote{\label{note: diglib} \scriptsize \urlstyle{tt}\url{https://diglib.natlib.lk/}}.

In the digital repository, we were able to organise all available content chronologically by issue date, allowing us to see publications printed dating back to 1800 CE. We carefully selected the book title, author name, identifier number, and collection name for each book from that point onward. This process required careful filtering, as most of the available content consisted of gazettes and police reports.

We identified 233 unique books printed between 1800 CE and 1955 CE in the \texttt{Natlib} digital repository (this is based on the issued date, not the written date). Of these, only 12 books were available for open access; we had to request access to the remainder. Our initial plan was to collect 100 sentences per year. To achieve this, we estimated that obtaining five pages of text from each book, excluding content pages and the preface, would provide us with more than 100 sentences, which amounts to approximately 1500 to 2000 word tokens, assuming there are about 15 to 20 words per sentence. Therefore, for the closed-access books, we requested five pages from each one. The process of obtaining access to these books was very difficult because most of them were part of the \texttt{Rare Books Collection}.

\subsection{Annotation by Written Date}
\label{subsec:annotate}

The issue date of the identified books was clearly stated in the digital repository at \texttt{Natlib}. However, this does not imply that the books were actually written during those specified dates. In fact, a book could have been written centuries earlier, while the printed version was released much later. 

Document dating has become extensively recognised in computational sociology and studies within digital humanities~\cite{ren-etal-2023-time,baledent-etal-2020-dating,hellwig-2020-dating}. When compared to other dating tasks, historical text dating is more complex due to the absence of explicit temporal indicators (such as time expressions) that aid in determining the date a document was written~\cite{toner2019language,baledent-etal-2020-dating,hellwig-2020-dating}.  It is clear that text dating, or the process of annotating the written date of a document, is an important task in diachronic studies~\cite{ansari2023diachronic, ren-etal-2023-time,favaro-etal-2022-towards}. 

Therefore, a comprehensive analysis was conducted to ensure that the written year of each book was accurately represented, ensuring that the resulting \texttt{SiDiaC} dataset accurately represents a proper diachronic corpus.

Upon the recommendation of experts in Sinhala linguistics, we identified a comprehensive book on Sinhala literature that claims to encompass literature information from its inception until 1994 CE~\cite{Sannasgala_2009}. This text served as the primary reference for the establishment of the respective date anchors, employing both time periods and specific years as outlined. The date ranges identified in~\citet{Sannasgala_2009} correspond either to the period during which the book was authored or to the time period in which the author lived.

The process of determining the written dates for the books became more complicated because some books in the dataset include commentaries and discourses on earlier works. In this version of the dataset, these cases are tied to the original earlier book's written date, as they contain both the information from the original book and its corresponding commentary (often written centuries prior, with extensive sections given as direct quotes without paraphrasing). 

\subsection{Challenges from Copyright Laws}

During the planning stage, one of the biggest challenges we faced was managing copyright issues. To address this, we conducted a thorough analysis of copyright laws in Sri Lanka, which are governed by the \texttt{Intellectual Property Act No. 36 of 2003}\footnote{\label{note: IPAct} \scriptsize \urlstyle{tt}\url{https://www.gov.lk/wordpress/wp-content/uploads/2015/03/IntellectualPropertyActNo.36of2003Sectionsr.pdf}}. 

According to this act, copyright in Sri Lanka is generally protected for the life of the author, plus an additional 70 years after their death. In cases where the author is unknown, copyright protection lasts for 70 years from the date of first publication. As a result, we focused on literature where the author passed away before 1955, as well as works by unknown authors that were published before 1955.

\subsection{Data Filtration}

\begin{figure}[!htbp]
    \centering
    \setlength{\fboxsep}{1pt} 
    \setlength{\fboxrule}{0.4pt} 
    \fbox{%
        \includegraphics[width=1.0\columnwidth]{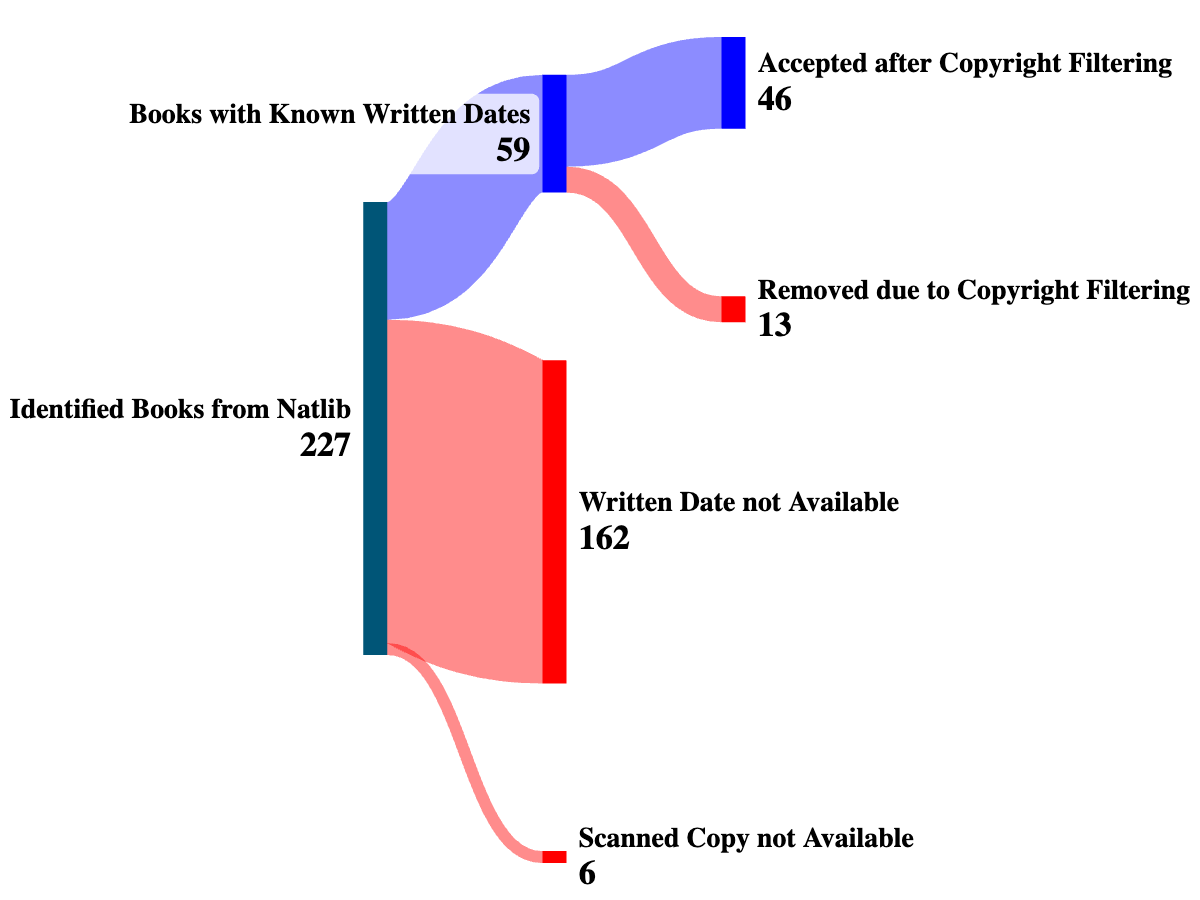}%
    }%
    \caption{Sequential Data Filtration Procedure}
    \label{fig:datafiltration}
\end{figure}

We initially identified 233 unique books, but after careful consideration of several factors, we ultimately selected only 46. Our selection process was influenced by the availability of scanned copies, the written dates of the works, and compliance with copyright laws as illustrated in Figure~\ref{fig:datafiltration}. 

Our first limitation was the availability of scanned copies at the digital repository of Natlib, which restricted us to 221 books. Additionally, we were only able to determine the written dates or periods for 65 of the 233 books. Taking both the availability of scanned copies and the accessibility of written dates into account, the number of selected records was reduced to 59. We then further refined this selection following the copyright law based filtering process we discussed, resulting in a final total of 46 books as presented in Table~\ref{tab:metadata_books} in Appendix~\ref{sec:app:lit}.

\subsection{Text Extraction using OCR}
\label{subsec:text_extract}

As the digital repository of Natlib shared the scanned copies of the requested books, we were forced to extract text information from the documents. Therefore, we selected an Optical Character Recognition (OCR) engine to get this task done. 


In the comparative study conducted by~\citet{jayatilleke2025zero}, five different OCR engines were analysed for text extraction using a synthetically created image-text dataset for Sinhala. Based on this study, we identified two standout OCR engines: \texttt{Google Document AI}\footnote{\scriptsize \urlstyle{tt}\url{https://cloud.google.com/document-ai/}} and \texttt{Surya}\footnote{\label{note:suryaocr}\scriptsize \urlstyle{tt}\url{https://github.com/VikParuchuri/surya}}, with \texttt{Surya} being reported to outperform all other systems compared. However, during our text extraction process, we found that, under realistic conditions (unlike the synthetic conditions used in the study), \texttt{Google Document AI} provided more accurate results as shown in Appendix~\ref{append:comp_sur_doc}.

\begin{figure}[!htbp]
    \centering
    \setlength{\fboxsep}{1pt} 
    \setlength{\fboxrule}{0.4pt} 
    \fbox{%
        \includegraphics[width=0.9\columnwidth]{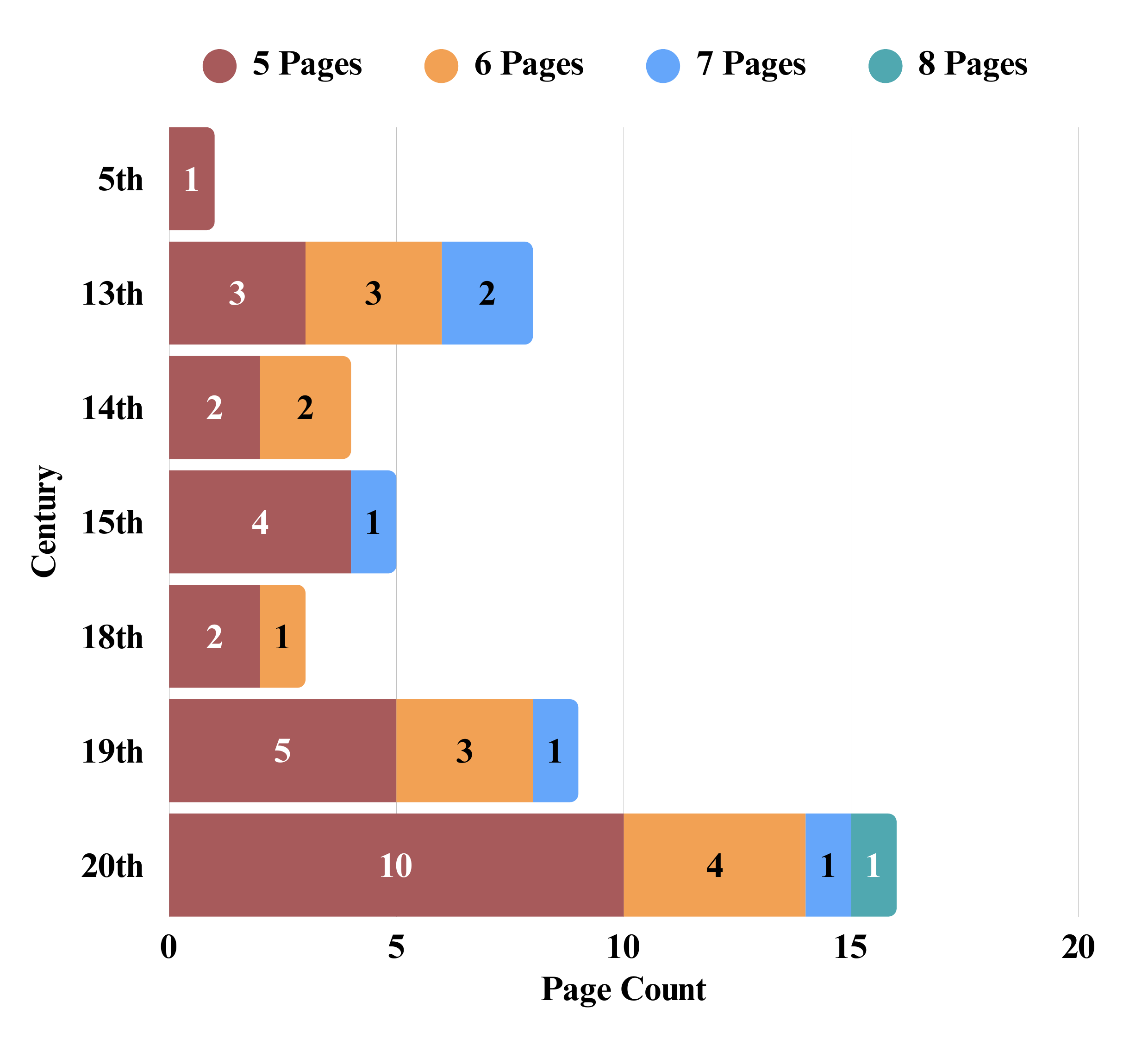}%
    }%
    \caption{Distribution of Page Counts of Scanned Copies}
    \label{fig:distpage}
\end{figure}

\texttt{Document AI} is a service provided by \texttt{Google Cloud Platform (GCP)}\footnote{\label{note:gcp}\scriptsize \urlstyle{tt}\url{https://cloud.google.com/}}. In this platform, we created a processor and utilised its Application Programming Interface (API) key to conduct OCR. The processor can handle a maximum of 15 pages at a time; however, this was not an issue since all of our scanned copies contained 5 to 8 pages, as shown in Figure~\ref{fig:distpage}. Throughout the procedure, we ensured that we obtained the model confidence for every page of each processed document. We then calculated the average confidence score, which is included in the metadata file of each book folder.

This OCR processor has demonstrated that it can perform text recognition that goes beyond simple extraction. It adapts effectively and generates words in modern Sinhala spelling while also taking into account morphology, where morphemes are formed accordingly, as explained in the section~\ref{subsec:ocr_perf}.

\subsection{Post-Processing Extracted Text}
\label{subsec:post-processing}

Although the OCR accuracy averaged 96.84\% across all documents, the formatting issues were significant enough to require manual adjustments. 
The Document AI's advanced performance helped to streamline manual post-processing tasks, significantly reducing the time required for that work.



The post-processing includes correcting the following text formatting issues:
\begin{itemize}
\setlength\itemsep{-0.1em}
\item \textbf{Spacing Errors}: This involves fixing incorrect or inconsistent spacing between words and sentences, such as missing spaces, uneven spaces and extra spaces.
\item \textbf{Multi-column Text}: This refers to texts containing errors where two columns were treated as a single column, resulting in entire horizontal lines being extracted without properly traversing each column separately.
\item ⁠\textbf{Misplaced Words/Phrases}: This addresses instances where words or phrases are out of order, leading to illogical text.
\item ⁠⁠⁠\textbf{Paragraph and Line Indentation}: This involves standardising the indentation of paragraphs and individual lines. This could mean adding consistent indents to new paragraphs (e.g., poem blocks) and removing incorrect indents.
\item \textbf{Removal of Seal Context}: This involves identifying and eliminating specific phrases or watermarks that represent seals or official stamps.

\item ⁠⁠\textbf{Page Number Removal}: This focuses on identifying and deleting page numbers that appear within the body of the text, as they are part of the document's structure rather than its content.
\end{itemize}

While language understanding was not a strict requirement for addressing these formatting-related factors, all manual post-processing procedures were carried out by the authors using a human-in-loop strategy~\cite{lamba2023exploring}. This approach involved correcting formatting errors within a single window that contained both page scans and editable transcripts~\cite{christy2017mass}. The authors responsible for these corrections are native Sinhala speakers. The post-processing steps applied, along with examples, are further discussed and illustrated in Appendix~\ref{append:post-processing}.

An important finding of this study was the presence of Pali, Sanskrit, and minimal English in certain records of \texttt{SiDiaC}. The inclusion of Pali and Sanskrit can be attributed to the fact that most historical texts are related to religion. Notably, both Pali and Sanskrit are written in the Sinhala script in all cases. In this study, we chose not to remove the content in these languages to avoid losing context.

\subsection{Creation of Metadata Files}

\begin{figure}[!htbp]
    \centering
    \setlength{\fboxsep}{1pt} 
    \setlength{\fboxrule}{0.4pt} 
    \fbox{%
        \includegraphics[width=0.9\columnwidth]{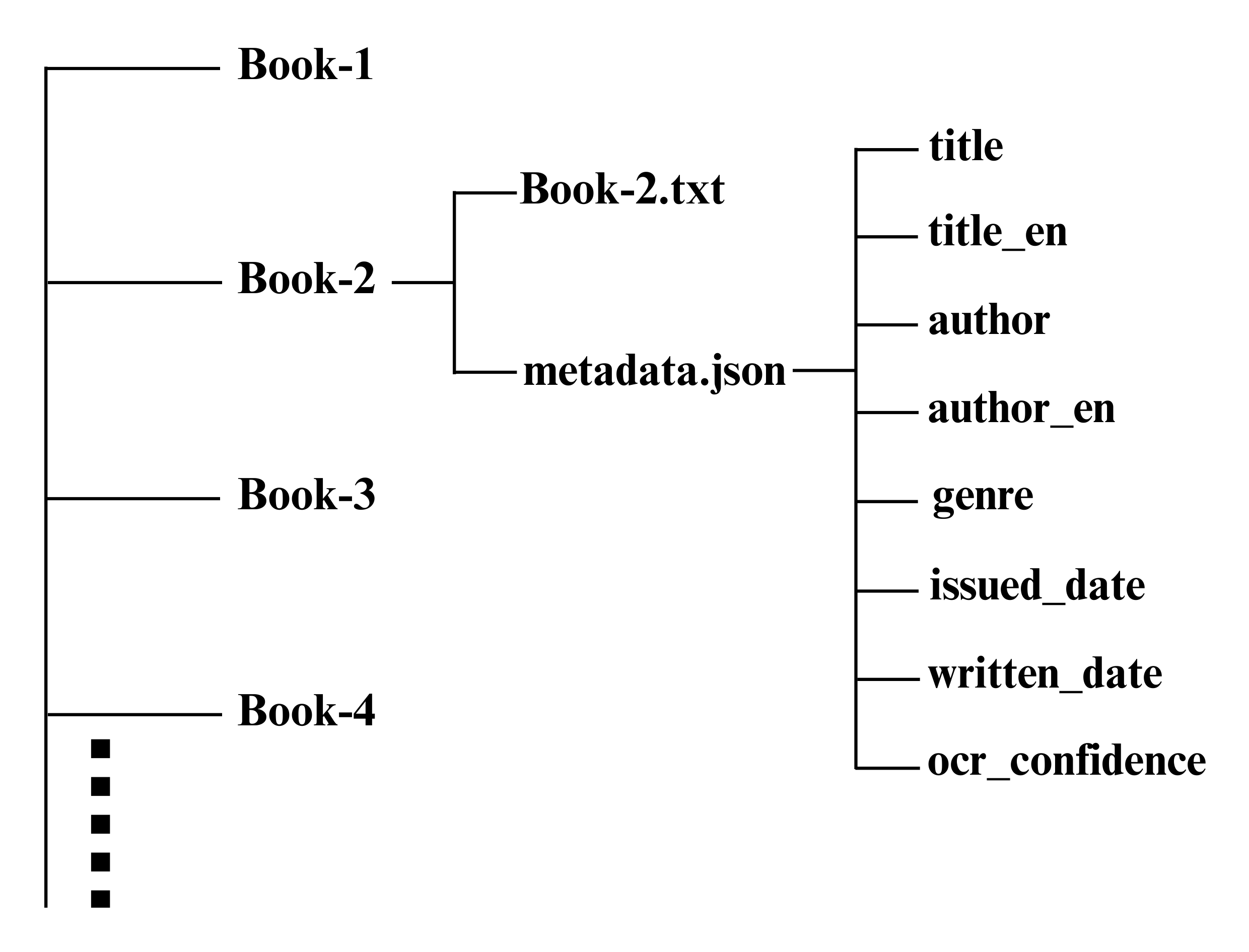}%
    }%
    \caption{Composition of the \texttt{SiDiaC} Corpus}
    \label{fig:datacomp}
\end{figure}

The dataset consisted of folders, each dedicated to a specific book. Within each folder, there is a text file along with a metadata file. The metadata files contain information such as the title and author names in both Sinhala and romanised forms, as well as the genre, issue date, written date, and the OCR confidence level for each particular book. Most of these information fields were identified through the \texttt{LatinlSE} corpus~\cite{mcgillivray2013tools}. Following the conventions of~\citet{davies2012expanding} and~\citet{rognvaldsson-etal-2012-icelandic}, we maintain a consistent metadata annotation method throughout the corpus without changing it across the centuries. The overall composition of the \texttt{SiDiaC} corpus, including folder and file level examples, is illustrated in Figure~\ref{fig:datacomp}.

The title of each book was consistently provided. When known, the authors' names were included; if the authors were unknown, they were labelled as \textit{unknown}. The issued date corresponds to the published year as listed in the digital repository of \texttt{Natlib}, while the written date was determined by referring to~\citet{Sannasgala_2009}, as explained in the section~\ref{subsec:annotate}. 

The genres of the books were selected based on the details provided by \citet{Sannasgala_2009}, as well as the content evaluated by authors who are native Sinhala speakers. The classification process occurs at two levels. The primary level is broad and divides the books into two categories: `Fiction' and `Non-Fiction.' The secondary level is more specific, categorising the content of the books into five distinct classes: religious, history, poetry, language, and medical. This approach was inspired by the methodologies followed in \texttt{IcePaHC} and \texttt{DIAKORP}~\cite{rognvaldsson-etal-2012-icelandic, kuvcera2015diakorp} corpora to ensure diverse genres. It is important to note that the first level of categorisation was applied to all documents, while the second level of categorisation was applied only to the books that fell into the selected specific categories. 

\begin{figure}[!htbp]
    \centering
    \setlength{\fboxsep}{1pt} 
    \setlength{\fboxrule}{0.4pt} 
    \fbox{%
        \includegraphics[width=0.8\columnwidth]{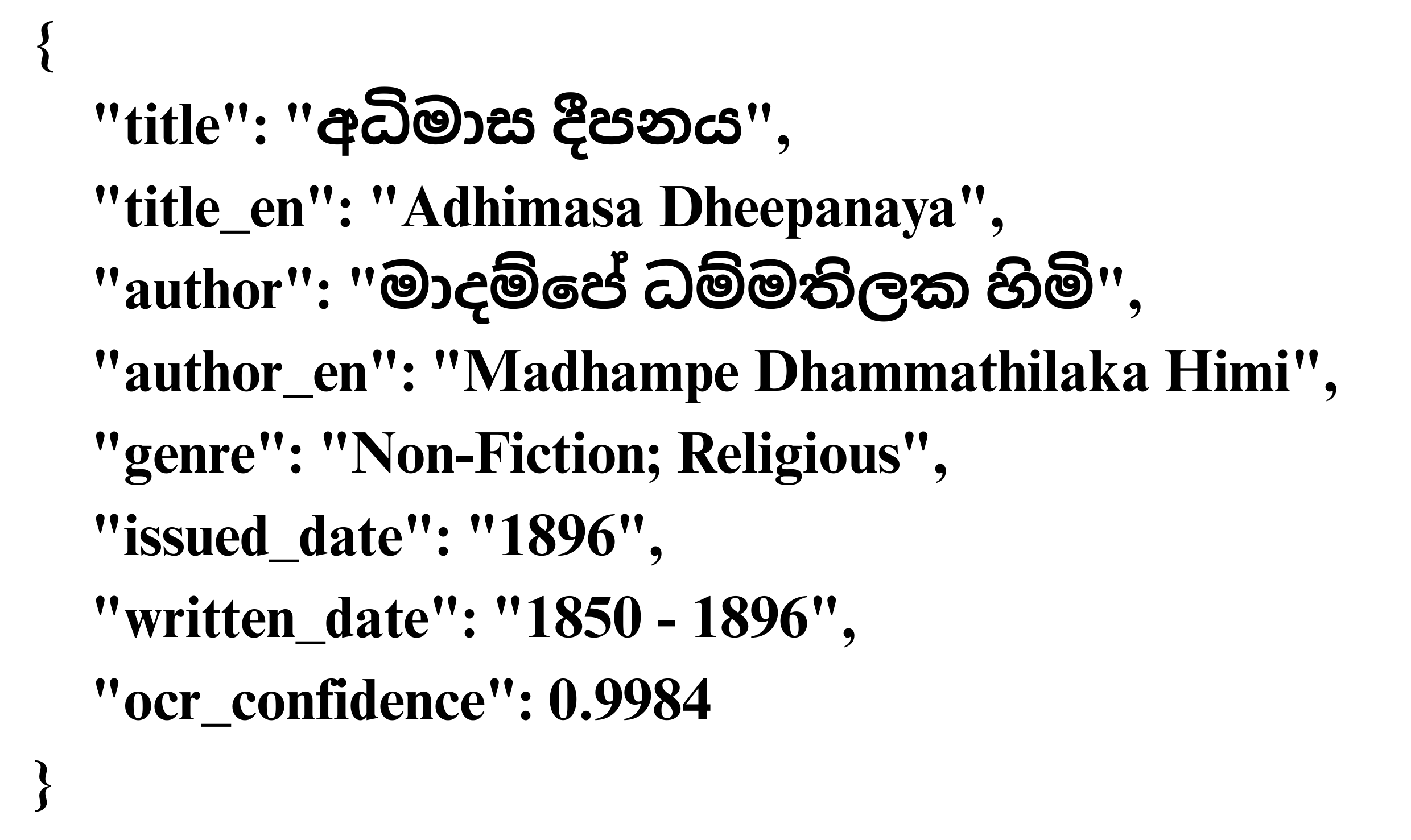}%
    }%
    \caption{An example of a metadata record in \texttt{SiDiaC}}
    \label{fig:exmet}
\end{figure}

Furthermore, the average OCR confidence per page is included, as mentioned in the section~\ref{subsec:text_extract}. In addition, we romanised the titles and author names for each book, which involved transliterating Sinhala content into the Latin alphabet. An example of a metadata record is shown in Figure~\ref{fig:exmet}.

\section{Analysis of \texttt{SiDiaC}}

\subsection{OCR Performance}
\label{subsec:ocr_perf}
\begin{figure}[!htbp]
    \centering
    \setlength{\fboxsep}{1pt} 
    \setlength{\fboxrule}{0.4pt} 
    \fbox{%
        \includegraphics[width=0.9\columnwidth]{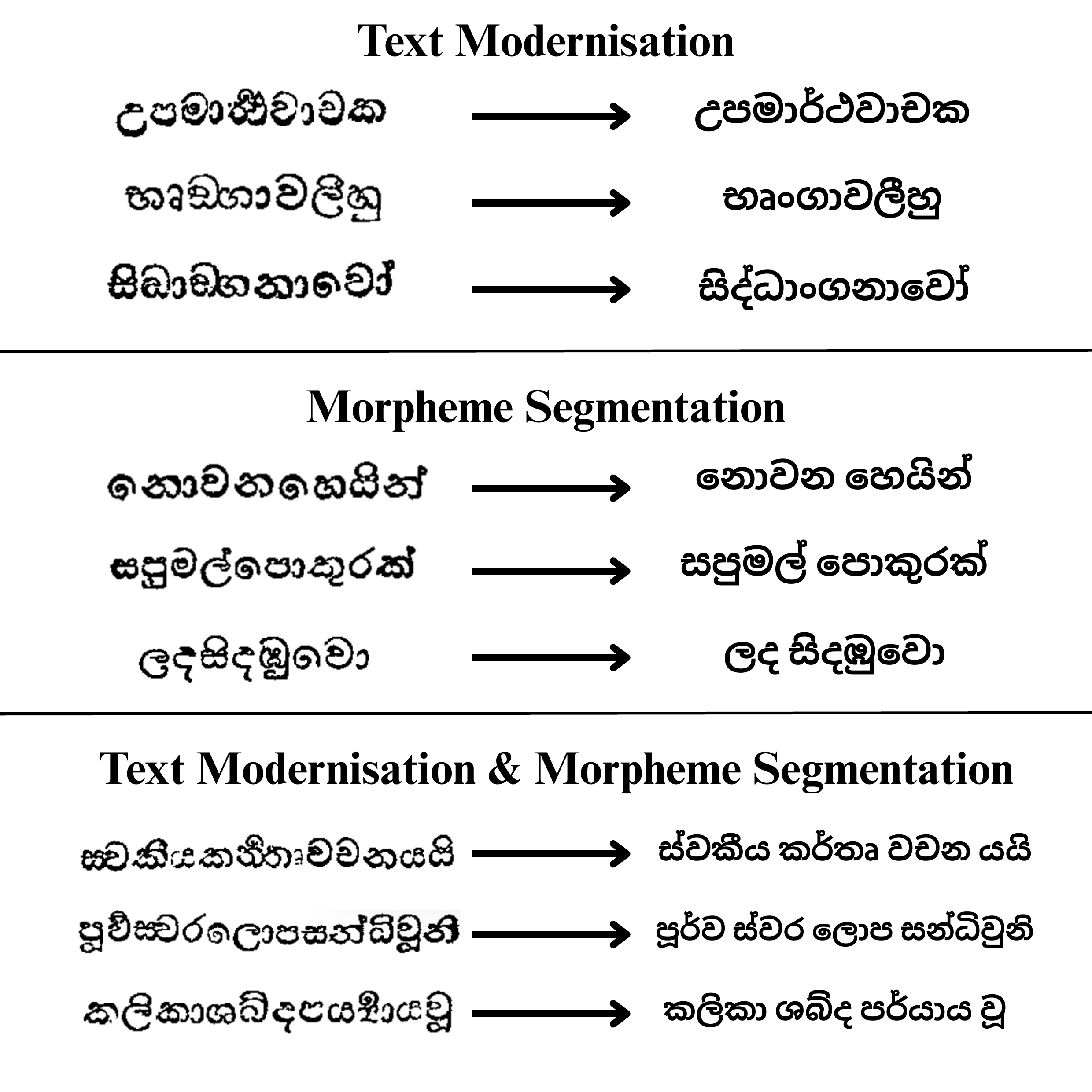}%
    }%
    \caption{Examples of Sinhala Text Modernisation and Morpheme Segmentation in \texttt{Document AI}}
    \label{fig:modern_seg}
\end{figure}

As previously noted, the performance of \texttt{Document AI} extended beyond simple text extraction as shown in Figure~\ref{fig:modern_seg}. These improvements helped to streamline manual post-processing tasks, significantly reducing the time required for that work.

Two significant additional steps were clearly observable to be performed by \texttt{Document AI}.

\begin{enumerate}
    \setlength\itemsep{-0.1em}
    \item ⁠⁠\textbf{Text Modernisation}: The historical development of the Sinhala language has resulted in various eras of syntax being linguistically equivalent but different in grapheme representation~\cite{nandasara2016bridging}. It is clear that \texttt{Document AI} adjusts to generate text in modern Sinhala syntax, ensuring a consistent and unified syntax throughout the dataset. Given that the change is only at the grapheme level, this does not violate the syntactic or semantic properties of the word.
    \item \textbf{Morpheme Segmentation}: In Historical Sinhala, certain words are combined without spaces, forming a closed compound~\cite{gaikwad2024Identification}. This phenomenon was accurately identified by \texttt{Document AI}, which effectively performs morpheme segmentation.
\end{enumerate}

Some characters in \texttt{SiDiaC} literature do not exist in the Sinhala Unicode. Therefore, mapping old characters to modern ones was an essential task performed by Document AI. Morpheme segmentation is crucial for maintaining consistency among words. If this issue is not addressed, combined multi-word expressions may be treated as unique words during the word embedding process, which can lead to significant differences in results when analysing semantic meaning.




\subsection{Evaluation of Metadata}

The dataset spans from the 5th to the 20th century CE, making it the longest continuous diachronic Sinhala corpus created to date. It covers many significant time periods, from the Anuradhapura era (377 BCE – 1017 CE) to just after Sri Lanka gained independence from Britain in 1948. This extensive timeframe allows for a representation of various changes in the language over the centuries.

\begin{table*}
  \centering
    \begin{tabularx}{\textwidth}{r|ZZ|ZZZZZ|Z}
  
    \hline
    & \multicolumn{2}{c|}{\textbf{Primary Category}} 
     & \multicolumn{5}{c|}{\textbf{Secondary Category}} &
     \multirow{2}{*}{     \textbf{Total}}\\
    \hhline{~-------~}
      & \multicolumn{1}{c}{\textbf{Non-Fiction}}& \multicolumn{1}{c|}{\textbf{Fiction}} & \multicolumn{1}{c}{\textbf{Religious}} & \multicolumn{1}{c}{\textbf{History}} & \multicolumn{1}{c}{\textbf{Poetry}} & \multicolumn{1}{c}{\textbf{Language}} & \multicolumn{1}{c|}{\textbf{Medical}} & \\
    \hline
    \textbf{5th} & 1 & 0 & 0 & 0 & 0 & 0 & 1 & 1       \\[2pt]
    \textbf{13th} & 7 & 1 & 5 & 0 & 1 & 2 & 0 & 8         \\[2pt]
    \textbf{14th} & 2 & 2 & 3 & 1 & 0 & 0 & 0 & 4       \\[2pt]
    \textbf{15th} & 1 & 4 & 2 & 0 & 3 & 0 & 0 & 5          \\[2pt]
    \textbf{18th} & 3 & 0 & 1 & 0 & 0 & 1 & 1 & 3          \\[2pt]
    \textbf{19th} & 6 & 3 & 2 & 2 & 3 & 2 & 0 & 9          \\[2pt]
    \textbf{20th} & 12 & 4 & 5 & 2 & 5 & 3 & 0 & *16        \\[2pt]\hline
    \textbf{Total} & 32 & 14 & 18 & 5 & 12 & 8 & 2 & 46\\\hline
   \end{tabularx}
  \caption{Distribution of Books Across Centuries and Genres. *The total count for the secondary category in the 20th century amounts to 15, while the overall number of books is 16. This discrepancy arises because the book `Hithopadhesha Sannaya', which offers advice, was not classified under any of the five secondary categories.}
  \label{tab:distbook}
\end{table*}
    
Assuming that the books with specified date ranges are attributed to the upper bound year, an analysis of the number of books per century was conducted, as illustrated in Table~\ref{tab:distbook}. The analysis reveals that the distribution of books in the corpus is heavily skewed toward the 20th century, with 28 out of 46 records originating after the 18th century. This trend may largely be attributed to the introduction of the printing press to Sri Lanka by the Dutch in 1737, which thereafter popularised book printing in the country~\cite{wickremasuriya1978beginnings,nandasara2016bridging}.

In the first level of genre classification between fiction and non-fiction, it is evident that there are more non-fiction books than fiction books in the corpus. At the second level of genre classification, religious texts and poetry dominate among the five categories. This predominance is largely due to the close relationship between Sinhala literary culture and Theravada Buddhism, which provided both subjects and a framework for preserving texts. Additionally, the influence of Sanskrit \textit{kavya} traditions and courtly patronage, which valued literary artistry and prestige, also played a significant role~\cite{hallisey2003works}.

The author of the book is known for 32 out of 46, while the remaining books are labelled as ``Unknown." Only three authors have published more than one book in this dataset: two authors each have two books, and one author has four.

The OCR confidence levels are extremely high, with an average of 96.84\% across all books and a minimum confidence score of 85.53\%. Despite these encouraging figures, it is clear that \texttt{Document AI} encountered various types of errors, which we largely addressed during the post-processing phase as discussed in Appendix~\ref{append:post-processing}. The accurate identification of characters and words likely contributes to these strong confidence scores; however, most errors appear to arise from the challenges presented by complex content formats.

\subsection{Evaluation of the Corpus}
\label{subsec:Evaluation}

\texttt{SiDiaC} consists of 58,027 word tokens that were filtered using \texttt{regex}, retaining only Sinhala and Latin characters, and subsequently tokenised by whitespace. The corpus contains 833 words in Latin script, which accounts for just 1.42\% of the entire dataset. Also, the complete dataset comprises 22,837 unique word tokens in Sinhala script, which accounts for 39.36\% unique word coverage of all words.
This total word token count, while not in the range of millions, such as the \texttt{COHA} corpus for English, comfortably passes the 53,000 token count of \texttt{FarPaHC} for Faroese, which is in the same language resource category as Sinhala according to~\citet{ranathunga-de-silva-2022-languages}.    

\begin{figure}[!htbp]
    \centering
    \setlength{\fboxsep}{1pt} 
    \setlength{\fboxrule}{0.4pt} 
    \fbox{%
        \includegraphics[width=0.9\columnwidth]{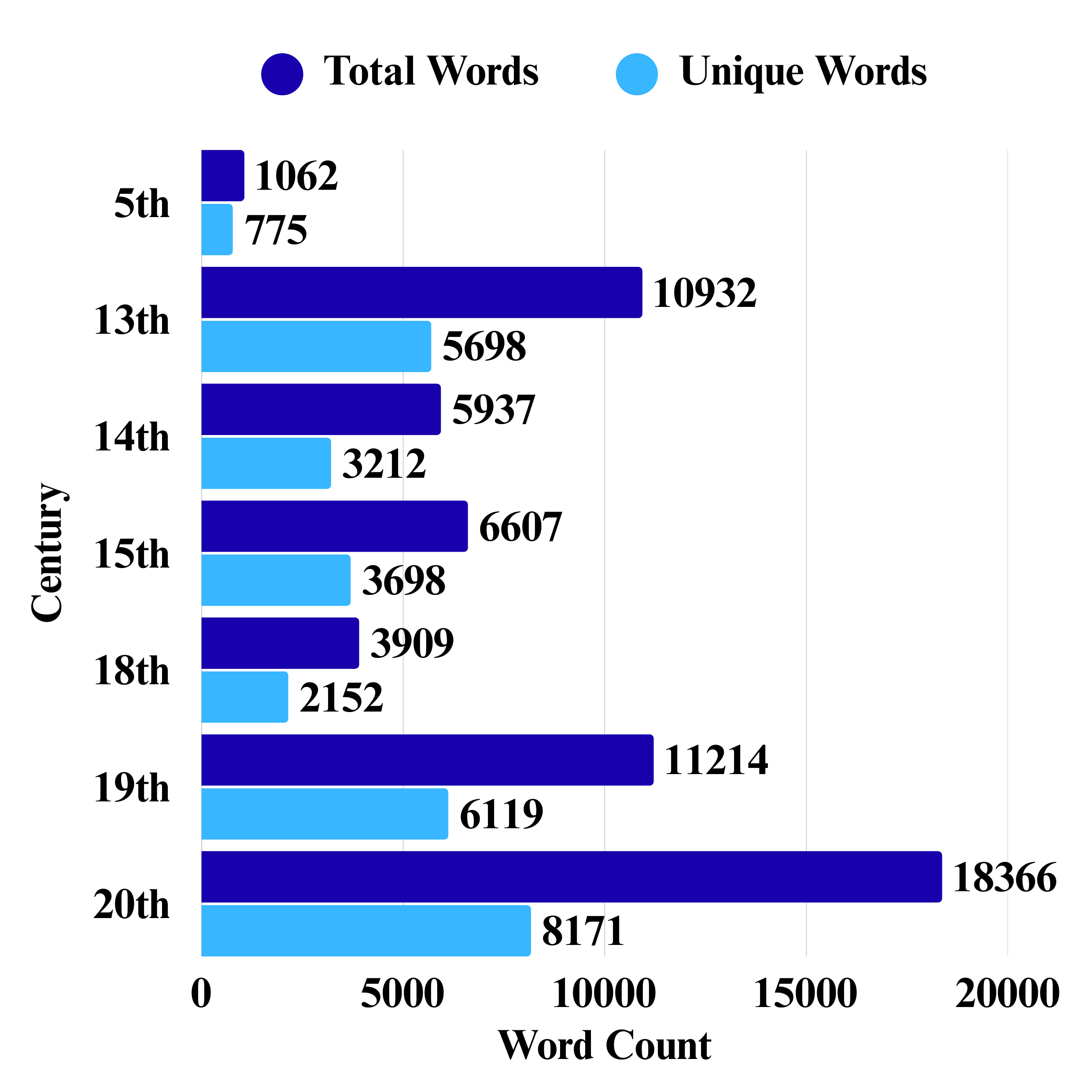}%
    }%
    \caption{Total vs Unique Word Counts per Century}
    \label{fig:wordco}
\end{figure}


In the 5th century, 72.98\% of words were unique, while the 13th century had 52.12\% unique words. The 14th century saw 54.1\%, the 15th century 55.97\%, and the 18th century 55.05\%. The 19th century featured 54.57\%, but by the 20th century, the percentage dropped to 44.49\%. As illustrated in Figure~\ref{fig:wordco}, generally a higher word count correlates with a lower percentage of unique words across the centuries.

Additionally, we conducted an analysis to identify the stopwords in the corpus by examining word tokens at the century level. Following the method described by~\citet{wijeratne2020sinhala} for their contemporary Sinhala corpus, we used a combination of word frequency analysis and manual vetting to identify appropriate stopwords from the corpora. To achieve this, we first counted the frequency of each unique word for each century and then converted these frequencies into z-scores.

\begin{equation}
  \label{eq:example}
  z_{w,c} = \frac{f_{w,c} - \mu_c}{\sigma_c}
\end{equation}

where, \( z_{w,c} \) is the z-score of the word \( w \) in the century \( c \). The term \( f_{w,c} \) refers to the frequency of the word \( w \) during that century, \( \mu_c \) denotes the mean frequency of all words in century \( c \), and \( \sigma_c \) indicates the standard deviation of the frequencies of all words in that century.


Next, we calculated \( -\infty < Z < 6.1027\) for a 99.80\% threshold. In the 20th century, we observed the highest number of word tokens, which was used to establish the threshold, assuming that it provided adequate coverage of stop words throughout the entire corpus. After sorting the words in the 20th century by z-score in descending order, we manually inspected the words to determine the ideal z-score as the upper limit.

Over the centuries, the number of identified stopwords has varied accordingly. In the 5th century, there were 47 stopwords, followed by 42 in the 13th century, 39 in the 14th century, 28 in the 15th century, 44 in the 18th century, 65 in the 19th century, and finally, 61 in the 20th century. 

The stopword that ranked highest across six of the seven centuries was `\raisebox{-0.5ex}{ 
    \includegraphics[height=1.3\fontcharht\font`\A]{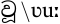} 
}'(was, became), while `\raisebox{-0.5ex}{ 
    \includegraphics[height=1.4\fontcharht\font`\A]{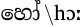} 
}'(or, either) ranked highest in the fifth century. Additionally, other frequently occurring stopwords include \raisebox{-0.5ex}{ 
    \includegraphics[height=1.4\fontcharht\font`\A]{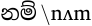} 
}(that, if), `\raisebox{-0.5ex}{
    \includegraphics[height=1.3\fontcharht\font`\A]{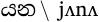} 
}'(that[named]), `\raisebox{-0.7ex}{
    \includegraphics[height=1.4\fontcharht\font`\A]{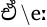} 
}'(that, those), `\raisebox{-0.5ex}{ 
    \includegraphics[height=1.4\fontcharht\font`\A]{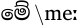} 
}'(this, these), `\raisebox{-0.5ex}{ 
    \includegraphics[height=1.3\fontcharht\font`\A]{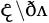} 
}'(or, interrogative particle), `\raisebox{-0.5ex}{ 
    \includegraphics[height=1.3\fontcharht\font`\A]{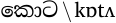} 
}'(while), `\raisebox{-0.5ex}{
    \includegraphics[height=1.4\fontcharht\font`\A]{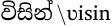} 
}'(by), and `\raisebox{-0.5ex}{ 
    \includegraphics[height=1.4\fontcharht\font`\A]{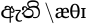} 
}'(be [exists]). However, it was observed that certain words, such as `\raisebox{-0.5ex}{
    \includegraphics[height=1.5\fontcharht\font`\A]{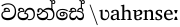} 
}'(honorific for a revered person) and `\raisebox{-0.5ex}{ 
    \includegraphics[height=1.4\fontcharht\font`\A]{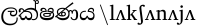}
}'(Quality) were also receiving high z-scores, potentially due to the corpus's strong connection to Buddhist literature. The methodology for identifying the top 48 stopwords in the entire corpus and their depiction is presented in Appendix~\ref{append:stopwords}.

\section{Future Work}

\texttt{SiDiaC}, being the first resource of its kind, paves the way for diachronic linguistic studies of the Sinhala language. We encourage researchers to explore areas such as lexical semantic change, tracking neologisms, grammatical change, historical language modelling, and corpus-based lexicography. Additionally, since the data is annotated by genre, it also allows for synchronic studies focusing on domain-based differences.

The corpus, as mentioned earlier, has comfortably surpassed the 53,000-token count of \texttt{FarPaHC} for Faroese, which falls into the same language resource category according to~\citet{ranathunga-de-silva-2022-languages}. However, the token count of \texttt{SiDiaC}, currently at 58,027, could certainly be increased by adding more literary works and additional pages from the existing 46 books. This enhancement would support more accurate research studies.

It is important to note that the OCR post-processing conducted in this study focuses only on formatting. However, as discussed in Appendix~\ref{append:word_char_errors}, there are clear issues present at the word and character levels that need to be meticulously addressed. Furthermore, as mentioned earlier, the corpus is code-mixed with Pali, Sanskrit, and English. This highlights the need for a processing step to identify and remove irrelevant text, ensuring that the corpus is entirely focused on Sinhala.

We also identified that books called `\raisebox{-0.35ex}{ 
    \includegraphics[height=1.5\fontcharht\font`\A]{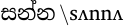} 
}' (meaning \textit{commentaries}) may include two dates: one for the original (quoted) text and another for the commentary. However, this study did not consider this phenomenon, and such instances were attributed to the well-known original version. Therefore, in future studies, it would be beneficial to identify the two dates and include the text sections originating from the different time periods at their correct positions in the corpus. 

The \texttt{SiDiaC} corpus did not undergo any lexical annotations during this study. Typically, the most recognised method for creating diachronic corpora involves parsing the entire corpus using POS tagging. However, this approach was not feasible with Sinhala POS taggers due to their limited performance. In future research, it would be highly beneficial to have the entire corpus parsed manually by Sinhala linguists who understand the evolution of language structure in Sinhala.

\section{Conclusion}

In this study, we introduced \texttt{SiDiaC}, a diachronic corpus of the Sinhala language. The corpus contains approximately 58k tokens, categorised into genres at two levels, and spans from the 5th century to the 20th century. This makes it the first diachronic Sinhala corpus ever created, which can serve as a foundational dataset for enhancing historical corpora in the Sinhala language. The entire process involved carefully identifying literature from the \texttt{Natlib} of Sri Lanka, followed by data filtering, date annotation, text extraction from PDF images, and post-processing. We also created metadata files containing important information about each book.

The complete corpus was thoroughly analysed, highlighting the powerful OCR performance of \texttt{Document AI} beyond simple text extraction. This was followed by a detailed evaluation of the dataset based on the metadata of all the books. Additionally, a comprehensive analysis was conducted at the word token level to ensure the identification of important findings within the corpus. 
Finally, we discussed potential future studies and approaches that could enhance the dataset, as well as the research opportunities that this corpus provides.

\section*{Limitations}

The creation of the corpus went through different types of limitations due to various challenges we faced. 


\paragraph{Literature Identification:}
While we recognised the \textit{Department of National Archives}\footnote{\label{note: natarch} \scriptsize \urlstyle{tt}\url{https://websnew.lithium.lk/archives/}} of Sri Lanka also as a credible source, data acquisition was conducted only from the \textit{National Library of Sri Lanka} due to permission constraints.

\paragraph{Data Filtration:} 
Out of the 221 scanned copies acquired, we were able to identify the written dates or periods for only 59 of them. The written dates of the books were annotated based on the lifespans of well-known authors, while the majority of the remainder were annotated relying heavily on the work by ~\citet{Sannasgala_2009}, which represents an overreliance on a single source.

\paragraph{Post-Processing after OCR:}  
Under this process, while corrections were initiated to address identified formatting issues, possible identification errors at the word or character level discussed in Appendix~\ref{append:word_char_errors} were not addressed.

\paragraph{Code-Mixed Data:}  
It was noted that the corpus contains code mixing of Pali, Sanskrit, and English languages, but the removal of text in these languages from the corpus has not been done.

\paragraph{Commentary Books:}  
The identified books primarily named with the term `\raisebox{-0.35ex}{ 
    \includegraphics[height=1.5\fontcharht\font`\A]{Figures/si_049.pdf} 
}' (meaning commentaries) will include two written dates for the original and the commentary. However, these instances were anchored to the well-known original version without removing the commentary.

\paragraph{Lexical Annotation:} 
Unlike the \texttt{LatinlSE}, \texttt{IcePAHC}, \texttt{COHA}, and \texttt{Google N-gram} corpora, which have undergone lexical annotations specifically for POS tagging, we were unable to conduct similar annotations due to the unavailability of Sinhala POS taggers~\cite{de2025survey}.

\section*{Acknowledgments}

The creation of the \texttt{SiDiaC} corpus was made possible through the valuable contributions of several individuals. We extend our sincere gratitude to \texttt{Padma Bandaranayake}, \textit{Director of the National Library \& Documentation Centre}, for her assistance with data acquisition. We also acknowledge \texttt{Uthpala Nimanthi} and \texttt{Charani Palangasinghe} for their efforts in the post-processing of the data, and the expertise of \texttt{Nalaka Jayasena}, a Sinhala Linguist, which was important in the identification of the book by \citet{Sannasgala_2009}. Finally, we would like to thank \texttt{Jayath de Silva}, \texttt{Savin Madapatha}, and \texttt{Thushan Bawantha} for their dedicated work on the written date annotation.


\bibliography{custom}

\appendix

\section{Utilised Literary Works}
\label{sec:app:lit}

During this study, we collected 46 literary works from the \texttt{Natlib} of Sri Lanka, including 32 by recognised authors and the rest by 'unknown' authors. \texttt{Munidhasa Kumarathunga} contributed four books, \texttt{Madhampe Dhammathilaka Himi} and \texttt{Hikkaduwe Sri Sumangala Himi} contributed two each, and the remaining authors each had one book, totalling 27 unique authors.

The metadata includes the title in Sinhala, the romanised title, the author's name, the romanised author's name, the genre, the issue date, the writing date, and the OCR confidence level. 


The complete metadata for each literary work used in this compilation of the dataset can be found in Table~\ref{tab:metadata_books} with the titles and authors' names presented in romanised Sinhala.
This diachronic spread ensures coverage of Sinhala evolution across medieval, pre-modern, and modern stages. Religious texts dominate, reflecting both preservation biases and the centrality of Buddhism in Sinhala literary culture.

\section{Comparison of \texttt{Document AI} \& \texttt{Surya}}
\label{append:comp_sur_doc}

The thorough quantitative analysis conducted by \citet{jayatilleke2025zero} indicates that \texttt{Surya} outperforms \texttt{Document AI} when evaluated on a synthetically created Sinhala dataset. However, during our text extraction process, we found that \texttt{Document AI} actually surpasses Surya. We believe this discrepancy stems from the synthetic data used in their study, which does not accurately reflect the challenges presented by real scanned documents.

Table~\ref{fig:surya_vs_document_AI} highlights three examples that clearly demonstrate why \texttt{Document AI} is the superior OCR engine. The errors indicated in red boxes for both systems demonstrate that \texttt{Document AI} excels in character identification, particularly with diacritics and similar-looking letters. Additionally, \texttt{Document AI} appears to be the only system effectively implementing morpheme segmentation, which is crucial for maintaining consistent word forms over time. Lastly, \texttt{Document AI}’s text modernisation feature provides another significant advantage, making it the ideal choice for integration into the OCR pipeline used in this study.

\section{Post-Processing Extracted Text}
\label{append:post-processing}

During this phase, we addressed six types of formatting issues. This careful task was carried out by the authors of this study, who are native Sinhala speakers.

Certain literary works contained unwanted text referred to as seal context, which did not relate to the books' actual content. As a result, this text was identified and removed from the book files in the dataset. Some examples of these seal contexts can be seen in Table~\ref{fig:seal_context}.

\begin{table}[h]
    \centering
    \setlength{\fboxsep}{1pt} 
    \setlength{\fboxrule}{0.4pt} 
    \fbox{%
        \includegraphics[width=0.9\columnwidth]{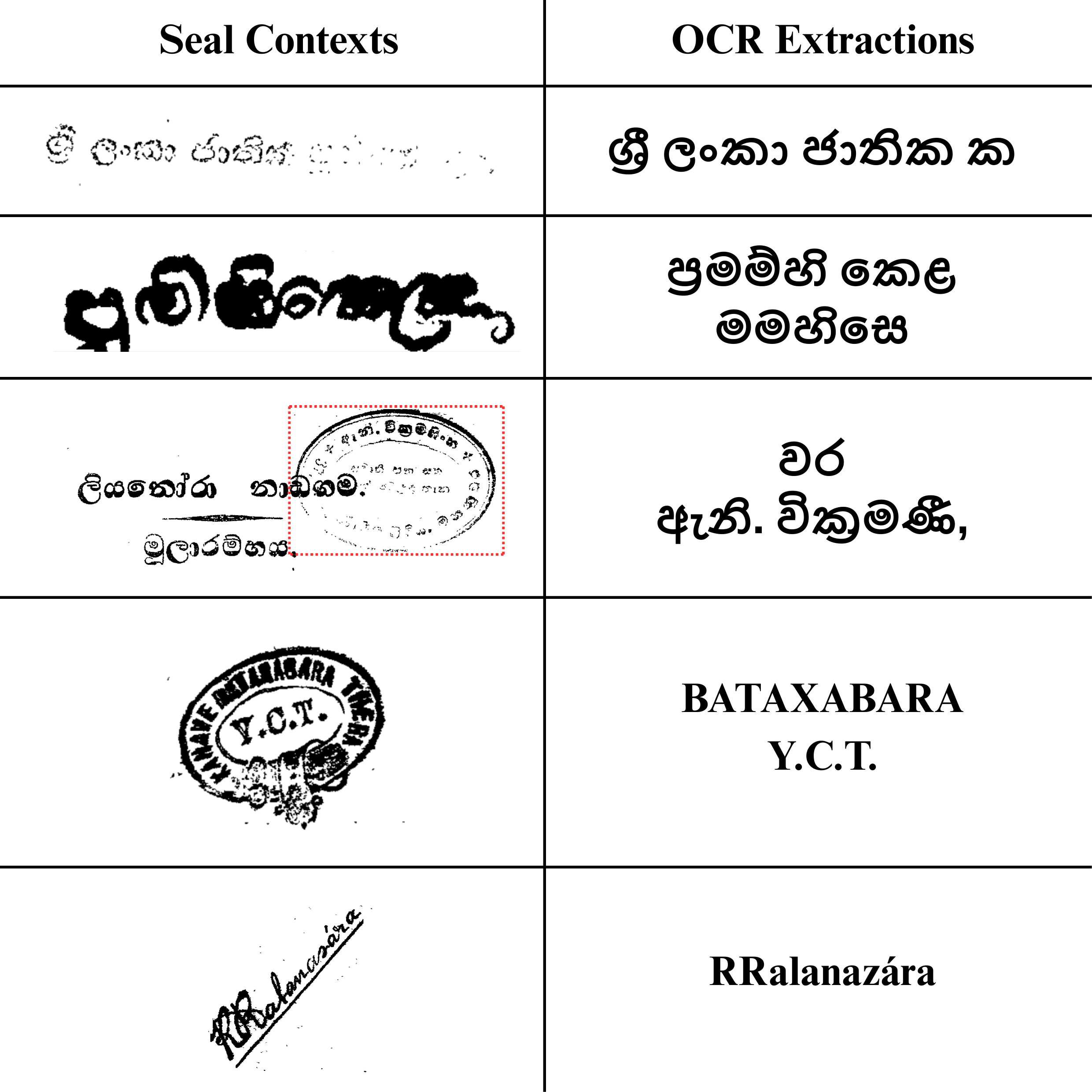}%
    }%
    \caption{Examples of seal contexts found in the PDF image documents and the corresponding extractions by \texttt{Document AI} that were removed during post-processing.}
    \label{fig:seal_context}
\end{table}

\begin{table*}[!htb]
\centering
\resizebox{\textwidth}{!}{
\begin{tabular}{l|c|c|c|c|c|c}
\hline
\textbf{Title} & \textbf{Author} & \multicolumn{2}{c|}{\textbf{Genre}} & \textbf{Issued Date} & \textbf{Written Date} & \textbf{OCR Confidence}$\uparrow$ \\
& & Primary & Secondary &&&\\
\hline
Adhimasa Dheepanaya & Madhampe Dhammathilaka Himi & Non-Fiction& Religious & 1896 & 1850 - 1896 & 0.9984\\
\hline
Adhimasa Winishchaya & Walikande Sri Sumangala Himi & Non-Fiction & Religious  & 1904 & 1850 - 1904 & 0.9971\\
\hline
Adhimasa Sangrahawa & Madhampe Dhammathilaka Himi & Non-Fiction & Religious  & 1903 & 1850 - 1903 & 0.9692\\
\hline
Anagathawanshaya: Methe Budu Siritha & \makecell{Watadhdhara Medhanandha Himi;\\Siri Parakumabahu;\\Wilgammula Sangaraja Himi}  & Fiction & Religious  & 1934 & 1325 - 1333 & 0.9992\\
\hline
Ashoka Shilalipi saha Prathimakarana Winishchaya & D. E. Wickramasuriya & Non-Fiction & History & 1919 & 1916 & 0.9989\\
\hline
Okandapala Sannaya hewath Balawathara Liyana Sanna & Don Andhris Silva & Non-Fiction & Language  & 1888 & 1760 - 1778 & 0.9884\\
\hline
Kavya Wajrayudhaya - Palamu Kotasa & Engalthina Kumari & Fiction & Poetry  & 1889 & 1825 - 1893 & 0.9254\\
\hline
Kavyashekaraya & Thotagamuwe Rahula Himi & Fiction & Poetry  & 1872 & 1408 - 1491 & 0.9813\\
\hline
Kudusika & Unknown & Non-Fiction & Poetry  & 1894 & 1270 - 1293 & 0.9988\\
\hline
Kusajathaka Wiwaranaya (Prathama Bagaya) & Munidhasa Kumarathunga & Non-Fiction & Religious  & 1932 & 1887 - 1932 & 0.9701\\
\hline
\makecell[l]{Gadaladeni Sannayai Prasidhdha wu\\ Balawathare Purana Wyakyanaya} & Hikkaduwe Sri Sumangala Himi &  Non-Fiction & Language  & 1877 & 1827 - 1911 & 0.9970\\
\hline
Jubili Warnanawa & John de Silva & Non-Fiction & Language  & 1887 & 1857 - 1922 & 0.9957\\
\hline
Dhaham Sarana & Unknown & Fiction & Religious  & 1931 & 1220 - 1293 & 0.9891\\
\hline
Dhaladha Pujawaliya & Unknown & Non-Fiction & History  & 1893 & 1325 - 1333 & 0.9978\\
\hline
Dhurwadhi Hardhaya Widharanaya & Sri Dhanudhdharacharya & Non-Fiction & Religious  & 1899 & 1854 - 1899 & 0.9919\\
\hline
Dhampiya Atuwa Gatapadaya & D.B. Jayathilaka & Non-Fiction & Religious  & 1932 & 1868 - 1932 & 0.9241\\
\hline
\makecell[l]{Dharma Pradheepikawa hewath\\Mahabodhiwansha Parikathawa} & Unknown & Non-Fiction & Religious  & 1906 & 1187 - 1225 & 0.9682\\
\hline
Dharmapradeepikawa & Gurulu Gomeen & Non-Fiction & Religious  & 1951 & 1187 - 1225 & 0.9786\\
\hline
Nikam Hakiyawa & Munidhasa Kumarathunga & Fiction & Poetry  & 1941 & 1887 - 1941 & 0.8932\\
\hline
Nikaya Sangrahaya hewath Shasanawatharaya & \makecell{Unknown} & Non-Fiction& Religious  & 1922 & 1390 & 0.9754\\
\hline
\makecell[l]{Nidhahase Manthraya} & \makecell{S Mahinda Himi} & Non-Fiction & Poetry & 1938 & 1901 - 1938 & 0.8997\\
\hline
\makecell[l]{Pansiya Panas Jathaka Potha} & \makecell{Unknown} & Fiction & Religious & 1881 & 1303 - 1333 & 0.9987\\
\hline
\makecell[l]{Parawi Sandheshaya} & \makecell{Unknown} & Fiction & Poetry & 1873 & 1430 - 1440 & 0.9902\\
\hline
\makecell[l]{Parani Gama} & \makecell{Galpatha Kemanandha Himi} & Non-Fiction & History & 1944 & 1944 & 0.9846\\
\hline
\makecell[l]{Budhdha Sikka hewath Kudu Sika} & \makecell{Unknown} & Non-Fiction & History & 1898 & 1270 - 1293 & 0.9699\\
\hline
\makecell[l]{Mage Malli} & \makecell{G. H Perera} & Non-Fiction & Poetry & 1938 & 1886 - 1938 & 0.8659\\
\hline
\makecell[l]{Mahawansa Teeka} & \makecell{Hikkaduwe Sri Sumangala Himi} & Non-Fiction & History & 1895 & 1827 - 1895 & 0.9978\\
\hline
\makecell[l]{Muwadew da Wiwaranaya} & \makecell{Munidhasa Kumarathunga} & Non-Fiction & Religious & 1949 & 1887 - 1944 & 0.8710\\
\hline
\makecell[l]{Moggalalayanawyakaranan} & \makecell{Moggallana Himi} & Non-Fiction & Language & 1890 & 1070 - 1232 & 0.9051\\
\hline
\makecell[l]{Moggallana Panchika Pradeepaya} & \makecell{Unknown} & Non-Fiction & Language & 1896 & 1070 - 1232 & 0.9918\\
\hline
\makecell[l]{Liyanora Nadagama} & \makecell{Unknown} & Fiction& Poetry & 1936 & 1852 - 1927 & 0.9935\\
\hline
\makecell[l]{Wibath Maldhama} & \makecell{Kirama Dhammarama Himi} & Non-Fiction & Language & 1906 & 1821 & 0.9986\\
\hline
\makecell[l]{Waidya Chinthamani Baishadhya Sangrahawa} & \makecell{Unknown} & Non-Fiction & Medical & 1909 & 1706 - 1739 & 0.9965\\
\hline
\makecell[l]{Wyakarana Wiwarana hewath\\Sinhala Bashawe Wyakaranaya} & \makecell{Munidhasa Kumarathunga} & Non-Fiction& Language & 1937 & 1887 - 1937 & 0.9029\\
\hline
\makecell[l]{Sadhdharma Rathnawaliya -Prathama Bagaya} & \makecell{Dharmasena Himi} & Non-Fiction& Religious & 1930 & 1220 - 1293 & 0.9962\\
\hline
\makecell[l]{Sanna sahitha Abhisambodhi Alankaraya} & \makecell{Waliwita Saranankara Sangaraja Himi} & Non-Fiction& Religious & 1897 & 1698 - 1778 & 0.9989\\
\hline
Sanna sahitha Salalihini Sandheshaya & \makecell{Unknown} & Fiction& Religious & 1859 & 1450 & 0.9909\\
\hline
\makecell[l]{Sanskrutha Shabdhamalawa\\hewath Sanskrutha Nama Waranagilla} & \makecell{Rathmalane Dharmaloka Himi} & Non-Fiction& Language & 1876 & 1828 - 1887 & 0.9671\\
\hline
\makecell[l]{Sarartha Sangrahawa: Prathama Bhagaya} & \makecell{Srimadh Budhdhadhasa Rajathuma} & Non-Fiction & Medical & 1904 & 398 - 426 & 0.9997\\
\hline
\makecell[l]{Sithiyam sahitha Mahiyangana Warnanawa} & \makecell{Unknown} & Fiction& Poetry & 1898 & 1878 & 0.9989\\
\hline
\makecell[l]{Sithiyam sahitha Sadhdharmalankaraya} & \makecell{Unknown} & Non-Fiction& Religious & 1954 & 1398 - 1410 & 0.9810\\
\hline
\makecell[l]{Sithiyam sahitha Siyabas Maldhama} & \makecell{Kirama Dhammanandha Himi} & Fiction & Poetry & 1894 & 1820 & 0.9256\\
\hline
\makecell[l]{Sithiyam sahitha Sinhala Mahawanshaya} & \makecell{D.H.S Abhayarathna} & Non-Fiction & History & 1922 & 1874 & 0.9549\\
\hline
\makecell[l]{Sinhala Wyakaranaya enam Sidath Sangarawa} & \makecell{Hikkaduwe Sri Sumangala Himi} & Non-Fiction &Language & 1884 & 1827 - 1911 & 0.9886\\
\hline
\makecell[l]{Hansa Sandheshaya} & \makecell{C.E. Godakumbure} & Fiction & Poetry & 1953 & 1457 - 1465 & 0.8553\\
\hline
\makecell[l]{Hithopadhesha Sannaya} & \makecell{Waligama Sri Sumangala Himi} & \makecell{Non-Fiction} & -& 1884 & 1825 - 1905 & 0.9871\\
\hline
\end{tabular}
}
\caption{\label{tab:metadata_books}
The metadata information for all the literature used in the creation of this dataset.
}
\end{table*}

\begin{table*}[h!tbp] 
    \centering

    \setlength{\fboxsep}{0.5pt} 
    \setlength{\fboxrule}{0.8pt}
    \fbox{%
    \includegraphics[width=0.93\textwidth]{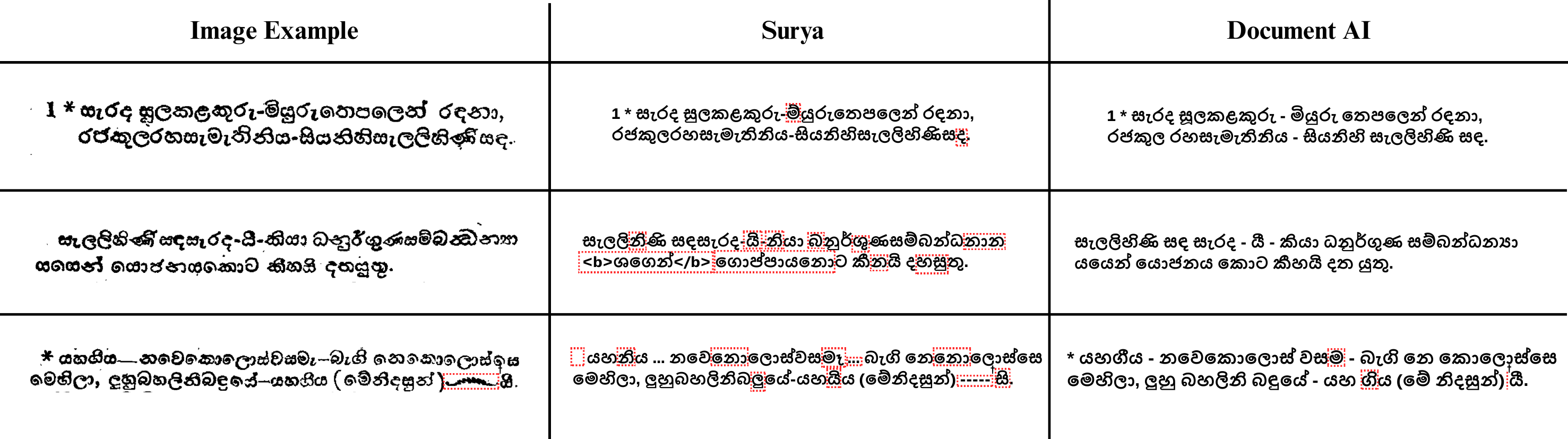} 
    }%
    \caption{Examples of sentences along with their corresponding text extractions from \texttt{Surya} and \texttt{Document AI} for comparison. Note that the characters and phrases highlighted in red boxes contain errors. $\dagger$ This character is known as `\raisebox{-0.3ex}{ 
    \includegraphics[height=2ex]{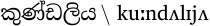} 
    }', a punctuation mark that indicates the end of a text or section in historical Sinhala.}
    \label{fig:surya_vs_document_AI}
\end{table*}

\begin{table*}[htbp!] 
    \centering

\setlength{\fboxsep}{0.5pt} 
    \setlength{\fboxrule}{0.8pt}
    \fbox{%
    \includegraphics[width=1.0\textwidth]{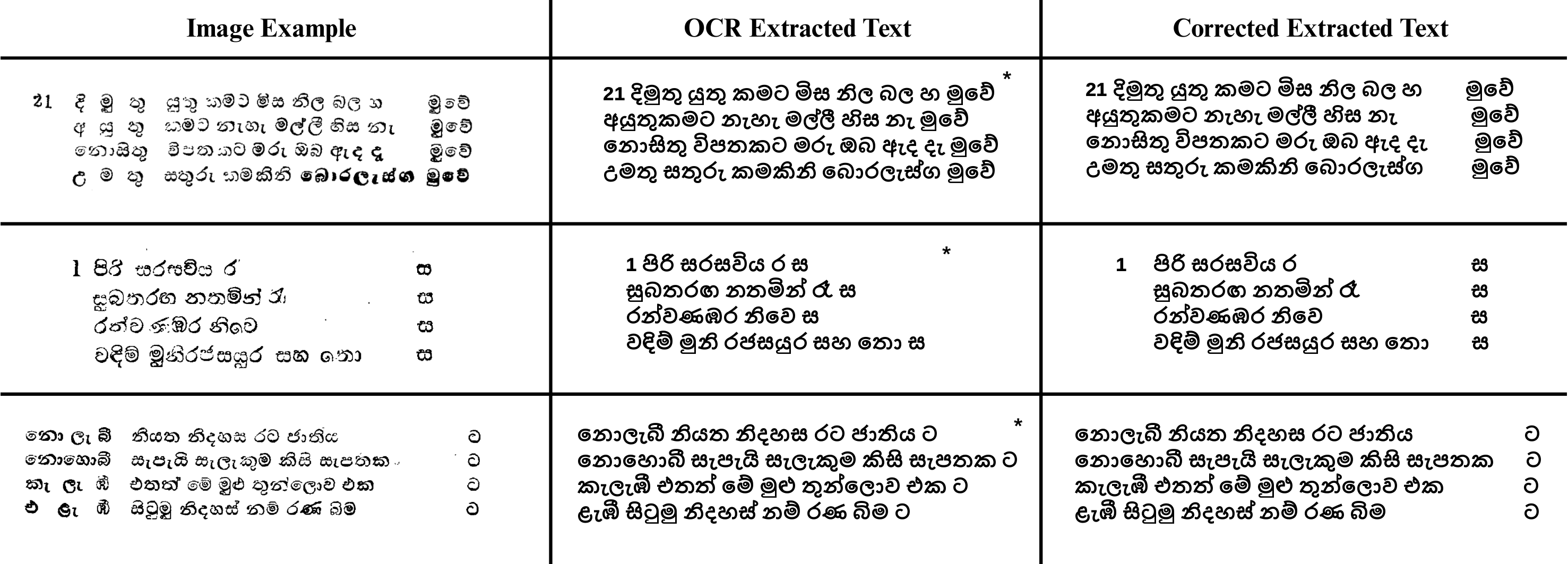} 
    }%
    \caption{Examples of spacing errors after OCR using \texttt{Document AI} on images and their corresponding corrections. *Note that the OCR extractions depicted were not exact; some final words were completely unidentified, which were added manually, and some had line breaks in awkward places.}
    \label{fig:spacing_errors}
\end{table*}

\begin{table*}[h!tbp] 
    \centering

    \setlength{\fboxsep}{0.5pt} 
    \setlength{\fboxrule}{0.8pt}
    \fbox{%
    \includegraphics[width=1.0\textwidth]{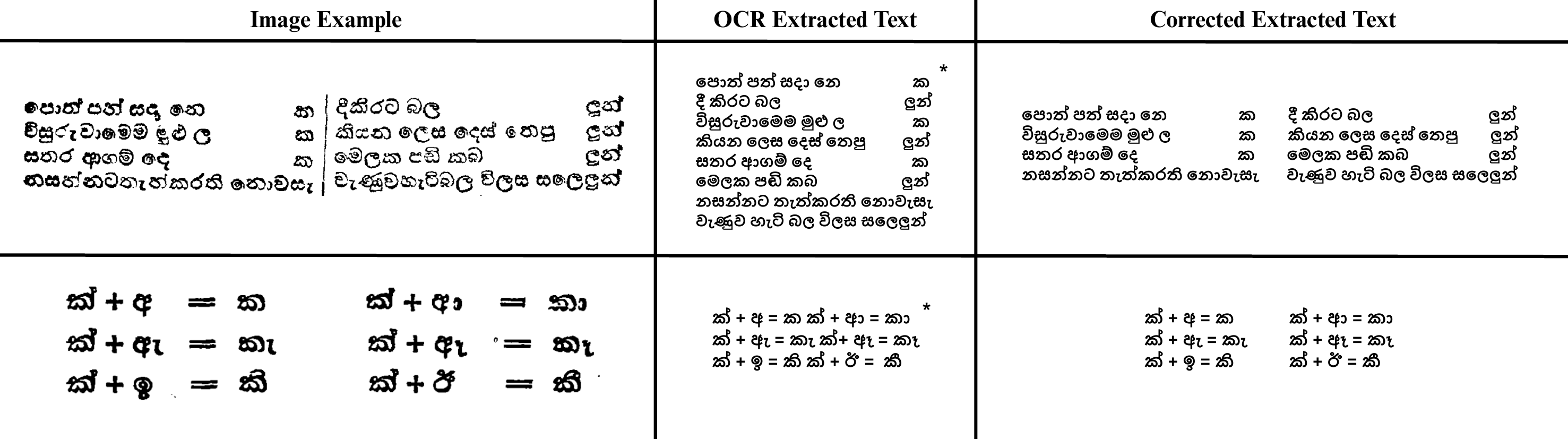} 
    }%
    \caption{Examples of errors in multi-column text after OCR using \texttt{Document AI} on images and their corresponding corrections. *Note that the OCR extractions shown here are not exact, as we could not fully represent an entire page that experienced this type of error in real case scenarios.}
    \label{fig:column_errors}
\end{table*}

\begin{table*}[h!tbp] 
    \centering

    \setlength{\fboxsep}{0.5pt} 
    \setlength{\fboxrule}{0.8pt}
    \fbox{%
    \includegraphics[width=1.0\textwidth]{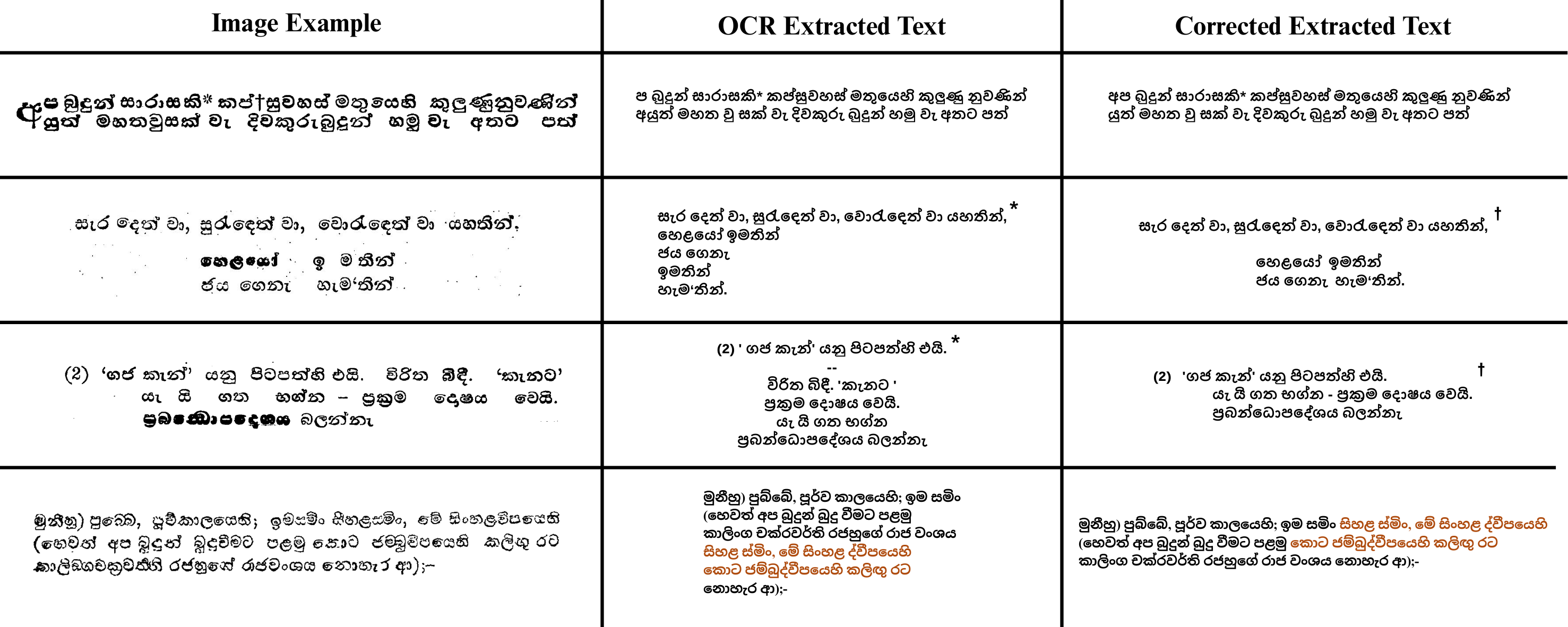} 
    }%
    \caption{Examples of misplaced words and phrases, along with $\dagger$errors in paragraph and line indentation that had occurred after using \texttt{Document AI} for OCR on images. Additionally, the corrections for these errors are provided. *Note that the OCR extractions displayed were not precise, as the errors were shown together rather than individually. The other errors were corrected to highlight the specific error being focused on.}
    \label{fig:misplaced_errors}
\end{table*}

\begin{table*}[h]
\centering
\resizebox{0.96\textwidth}{!}{
\begin{tabular}{c|c|c|c|c|c|c|c|c}
\hline
\textbf{Stopword} & \textbf{Meaning [in Context]}&  \textbf{5th} & \textbf{13th} & \textbf{14th} & \textbf{15th} & \textbf{18th} & \textbf{19th} & \textbf{20th} \\
\hline
\raisebox{-0.5ex}{ 
    \includegraphics[height=1.3\fontcharht\font`\A]{Figures/si_001.pdf} 
} & [that which came to] be & X & X & X & X & X & X & X \\
\hline
\raisebox{-0.5ex}{ 
    \includegraphics[height=1.4\fontcharht\font`\A]{Figures/si_003.pdf} 
} & that, if & X & X & X & X & X & X & X \\
\hline
\raisebox{-0.5ex}{ 
    \includegraphics[height=1.2\fontcharht\font`\A]{Figures/si_004.pdf} 
} & that [named] & X & X & X & X & X & X & X \\
\hline
\raisebox{-0.5ex}{ 
    \includegraphics[height=1.5\fontcharht\font`\A]{Figures/si_005.pdf} 
} & that, those &  & X & X &  & X & X & X \\
\hline
\raisebox{-0.5ex}{ 
    \includegraphics[height=1.4\fontcharht\font`\A]{Figures/si_006.pdf} 
} & this, these & X & X & X & X & X & X & X \\
\hline
\raisebox{-0.5ex}{ 
    \includegraphics[height=1.3\fontcharht\font`\A]{Figures/si_007.pdf} 
} & or, interrogative particle  & X & X &  & X & X & X & X \\
\hline
\raisebox{-0.5ex}{ 
    \includegraphics[height=1.2\fontcharht\font`\A]{Figures/si_008.pdf} 
} & while &  & X & X & X & X & X & X \\
\hline
\raisebox{-0.7ex}{ 
    \includegraphics[height=1.4\fontcharht\font`\A]{Figures/si_009.pdf} 
} & by & X & X & X &  & X & X & X \\
\hline
\raisebox{-0.5ex}{ 
    \includegraphics[height=1.3\fontcharht\font`\A]{Figures/si_010.pdf} 
} & be [exists] &  & X & X & X & X & X & X \\
\hline
\raisebox{-0.5ex}{ 
    \includegraphics[height=1.2\fontcharht\font`\A]{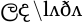} 
} & that [which was] & X & X & X & X &  & X & X \\
\hline
\raisebox{-0.5ex}{ 
    \includegraphics[height=1.4\fontcharht\font`\A]{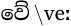} 
} & be [affirmed] &  & X &  & X & X &  & X \\
\hline
\raisebox{-0.5ex}{ 
    \includegraphics[height=1.4\fontcharht\font`\A]{Figures/si_002.pdf} 
} & or & X & X &  &  &  &  & X \\
\hline
\raisebox{-0.5ex}{ 
    \includegraphics[height=1.2\fontcharht\font`\A]{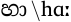} 
} & and &  & X & X & X & X & X & X \\
\hline
\raisebox{-0.5ex}{ 
    \includegraphics[height=1.2\fontcharht\font`\A]{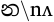} 
} & by &  & X &  &  &  & X & X \\
\hline
\raisebox{-0.5ex}{ 
    \includegraphics[height=1.2\fontcharht\font`\A]{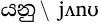} 
} & is [so named] &  & X &  &  & X & X & X \\
\hline
\raisebox{-0.5ex}{ 
    \includegraphics[height=1.2\fontcharht\font`\A]{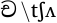} 
} & with, by &  & X &  &  &  & X & X \\
\hline
\raisebox{-0.5ex}{ 
    \includegraphics[height=1.2\fontcharht\font`\A]{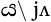} 
} & be [affirmed] &  & X &  &  &  & X & X \\
\hline
\raisebox{-0.5ex}{ 
    \includegraphics[height=1.3\fontcharht\font`\A]{Figures/si_011.pdf} 
}* & Honourable [suffix]
&  & X & X &  &  & X & \\
\hline
\raisebox{-0.5ex}{ 
    \includegraphics[height=1.2\fontcharht\font`\A]{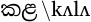} 
} & do [end of action] &  &  &  &  & X & X & X \\
\hline
\raisebox{-0.5ex}{ 
    \includegraphics[height=1.2\fontcharht\font`\A]{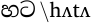} 
} & to & X &  &  &  & X &  &  \\
\hline
\raisebox{-0.5ex}{ 
    \includegraphics[height=1.3\fontcharht\font`\A]{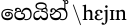} 
} & because &  & X & X &  & X &  & X \\
\hline
\raisebox{-0.5ex}{ 
    \includegraphics[height=1.3\fontcharht\font`\A]{Figures/si_012.pdf} 
}* & Quality & X &  &  &  &  &  &  \\
\hline
\raisebox{-0.5ex}{ 
    \includegraphics[height=1.2\fontcharht\font`\A]{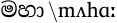} 
}* & Great [Prefix] &  & X & X &  &  & X & X \\
\hline
\raisebox{-0.5ex}{ 
    \includegraphics[height=1.2\fontcharht\font`\A]{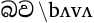} 
} & [the fact] that &  &  &  &  & X & X & X \\
\hline
\raisebox{-0.5ex}{ 
    \includegraphics[height=1.2\fontcharht\font`\A]{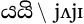} 
} & so [called] &  &  & X &  & X & X & X \\
\hline
\raisebox{-0.5ex}{ 
    \includegraphics[height=1.2\fontcharht\font`\A]{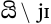} 
} & be [reported] &  & X & X &  &  &  & X \\
\hline
\raisebox{-0.5ex}{ 
    \includegraphics[height=1.2\fontcharht\font`\A]{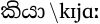} 
} & is [reported] &  &  &  &  & X & X & X \\
\hline
\raisebox{-0.5ex}{ 
    \includegraphics[height=1.2\fontcharht\font`\A]{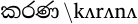} 
} & by [means of] &  &  &  & X &  & X & X \\
\hline
\raisebox{-0.5ex}{ 
    \includegraphics[height=1.3\fontcharht\font`\A]{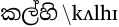} 
} & while, when &  & X & X &  & X &  &  \\
\hline
\raisebox{-0.5ex}{ 
    \includegraphics[height=1.4\fontcharht\font`\A]{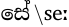} 
} & like [in manner of] &  & X &  &  &  &  & X \\
\hline
\raisebox{-0.5ex}{ 
    \includegraphics[height=1.2\fontcharht\font`\A]{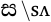} 
} & [poetic suffix] &  &  &  & X &  & X &  \\
\hline
\raisebox{-0.5ex}{ 
    \includegraphics[height=1.2\fontcharht\font`\A]{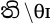} 
} & be [affirmed] &  & X &  &  &  & X &  \\
\hline
\raisebox{-0.5ex}{ 
    \includegraphics[height=1.2\fontcharht\font`\A]{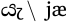} 
} & be [reported] &  & X &  &  &  &  & X \\
\hline
\raisebox{-0.5ex}{ 
    \includegraphics[height=1.2\fontcharht\font`\A]{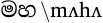} 
}* & Great [Prefix] &  & X & X &  &  &  &  \\
\hline
\raisebox{-0.5ex}{ 
    \includegraphics[height=1.25\fontcharht\font`\A]{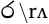} 
} & [poetic suffix] &  &  &  & X &  &  & X \\
\hline
\raisebox{-0.5ex}{ 
    \includegraphics[height=1.2\fontcharht\font`\A]{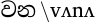} 
} & be [known to exist] &  & X &  &  &  &  & X \\
\hline
\raisebox{-0.5ex}{ 
    \includegraphics[height=1.3\fontcharht\font`\A]{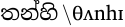} 
} & at, after [so declared] &  & X &  &  & X &  &  \\
\hline
\raisebox{-0.5ex}{ 
    \includegraphics[height=1.2\fontcharht\font`\A]{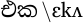} 
} & a/the/one &  &  &  &  &  & X & X \\
\hline
\raisebox{-0.5ex}{ 
    \includegraphics[height=1.15\fontcharht\font`\A]{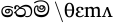} 
} & him & X &  &  &  & X &  &  \\
\hline
\raisebox{-0.5ex}{ 
    \includegraphics[height=1.15\fontcharht\font`\A]{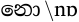} 
} & not &  & X &  &  &  &  & X \\
\hline
\raisebox{-0.5ex}{ 
    \includegraphics[height=1.3\fontcharht\font`\A]{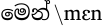} 
} & also [in similar manner] &  &  &  & X &  &  & X \\
\hline
\raisebox{-0.5ex}{ 
    \includegraphics[height=1.2\fontcharht\font`\A]{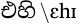} 
} & therein &  &  & X &  &  &  & X \\
\hline
\raisebox{-0.5ex}{ 
    \includegraphics[height=1.2\fontcharht\font`\A]{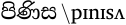} 
} & for &  &  &  &  &  & X &  \\
\hline
\raisebox{-0.5ex}{ 
    \includegraphics[height=1.3\fontcharht\font`\A]{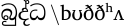} 
}* & Buddha &  &  & X &  & X &  &  \\
\hline
\raisebox{-0.5ex}{ 
    \includegraphics[height=1.2\fontcharht\font`\A]{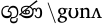} 
}* & Quality &  &  &  & X &  &  & X \\
\hline
\raisebox{-0.5ex}{ 
    \includegraphics[height=1.2\fontcharht\font`\A]{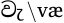} 
} & [after] be &  & X &  &  &  &  & X \\
\hline
\raisebox{-0.5ex}{ 
    \includegraphics[height=1.2\fontcharht\font`\A]{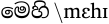} 
} & herein &  &  &  &  &  &  & X \\
\hline
\raisebox{-0.5ex}{ 
    \includegraphics[height=1.2\fontcharht\font`\A]{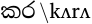} 
} & do [end of action] &  &  &  & X &  &  & X \\
\hline
\end{tabular}
}
\caption{\label{tab:stopwords}\footnotesize
The top 48 stopwords and their presence in each century after applying the threshold \( -\infty < Z < 6.1027\). *Note that some words, which are not technically stopwords, were included in this analysis. This may have occurred due to the limited availability of literary works in certain centuries and a bias toward Theravada Buddhism in the selected literature.
}
\end{table*}

It was notable to see errors in spacing in poems, especially when the last word or letter of a word is separated by multiple spaces that were not correctly detected by \texttt{Document AI}, as shown in Table~\ref{fig:spacing_errors}. Although correcting these errors may not hold significant semantic importance, accurately replicating the structure of the poems is crucial for preserving the original form of the books in case any downstream task using our dataset requires it. But for any task that does not require the original structure and would work on the lexical or semantic properties of the writing, the \texttt{Document AI} output is adequate. 
There were also multi-column texts, particularly evident in poetry books such as \textit{`Kavya Wajrayudhaya - Palamu Kotasa'}. These texts contained errors where two columns were treated as a single column, resulting in entire horizontal lines being extracted without properly traversing each column, as shown in Table~\ref{fig:column_errors}. To address these errors, corrections were made by mimicking the original structure of the books to preserve their intended format. 

The corrections provided in the table demonstrate that OCR outputs can be restructured faithfully only through column-aware preprocessing (for example, layout analysis, region detection, or image segmentation).

The misplaced words and phrases appeared mul-
\FloatBarrier

\noindent
-tiple times due to varying spacing and text positioning styles used in different books. 
Unlike simple character substitutions, these errors dislocate words or phrases from their expected syntactic and semantic slots. For example, text fragments are merged across lines, while key markers or paragraph boundaries vanish.
This was particularly noticeable at the beginning of the first paragraph in texts with drop caps. Similarly, in poetry, irregular line breaks caused by extra spaces before the last letter or word were among these issues. 
A misplaced suffix or dislocated phrase could invert rhetorical emphasis or obscure reference chains. Importantly, these errors also impact computational parsing, where algorithms expecting consistent lineation and phrase boundaries will misinterpret discourse structure. 
Another one of the most common issues addressed during post-processing was the indentation errors in paragraphs and lines. These errors were often found in the first line of paragraphs, with certain poems being misaligned, and page headings being centred incorrectly. A few examples of these issues are presented in Table~\ref{fig:misplaced_errors}.

\texttt{Document AI} also captured meta information on the physical book, such as page numbers. During the text extraction, these were removed from the books because they do not add any value to the presented corpus. These numbers were typically placed at the top or bottom, and they were centred or right-aligned in different literary works.

\section{Analysis of Stop Words in \texttt{SiDiaC}}
\label{append:stopwords}

The stopword analysis was conducted using the z-score calculation, as detailed in section~\ref{subsec:Evaluation}. During this analysis, we identified a union set of words that exceeded the established threshold, resulting in a total of 194 unique words. The words were sorted by their average z-score for each century in descending order. The list of words that were above the set threshold was cross-checked with the union set, illustrating their availability in each century. The top 48 words with the highest mean z-scores are displayed in Table~\ref{tab:stopwords}.

The continued inclusion of particles and suffixes shows continuity of core grammatical function words. Their presence across the 5th–20th centuries suggests remarkable diachronic stability in Sinhala morphosyntactic scaffolding. However, the table also reveals anomalies: certain lexical items not traditionally classified as stopwords appear in the stopword list. This distortion is likely due to corpus composition (religious texts with Buddhist themes dominate some centuries, inflating the relative frequency of doctrinal terms). Another observation is the persistence of poetic suffixes in older centuries, gradually tapering in modern texts.

\noindent
This diachronic shift may point to a movement away from poetry and towards prose. The presence/absence patterns also reveal data sparsity in some centuries, as marked by gaps where certain stopwords do not appear due to limited surviving texts. 
To wit, it is as interesting (mayhap more) to note what is missing in the 20th century, given that it seems to count a majority of the candidate words among its stopwords. Note how the archaic forms of honorifics and some direct references to Buddhism have dropped out of the list. 

In earlier centuries, the preponderance of monastic authorship and the dominance of canonical or exegetical works meant that words indexing reverence and religious entities were unavoidable high-frequency items.
Their disappearance, or at least their reduced prominence, in the most recent century may be reflecting how the subjects covered in the text have shifted from esoteric religious communication to comparatively more secular discourse.    
Equally important is the fact that the persistence of other grammatical function words across all centuries stands in stark contrast to this attrition of religiously marked lexemes. This points to a kind of lexical stratification: the unmarked syntactic scaffolding of Sinhala remains stable over time (only being replaced by synonyms when they do), while the culturally bound vocabulary tied to ritual, doctrine, or honorific practice is more vulnerable to historical change.

\section{Word \& Character Level Errors}
\label{append:word_char_errors}

During the post-processing of extracted text from scanned PDF files, we carefully conducted formatting level corrections as mentioned in section~\ref{subsec:post-processing}. However, these adjustments did not resolve all the necessary corrections at the word and character levels.

\begin{table}[!htbp]
    \centering
    \setlength{\fboxsep}{1pt} 
    \setlength{\fboxrule}{0.4pt} 
    \fbox{%
        \includegraphics[width=0.9\columnwidth]{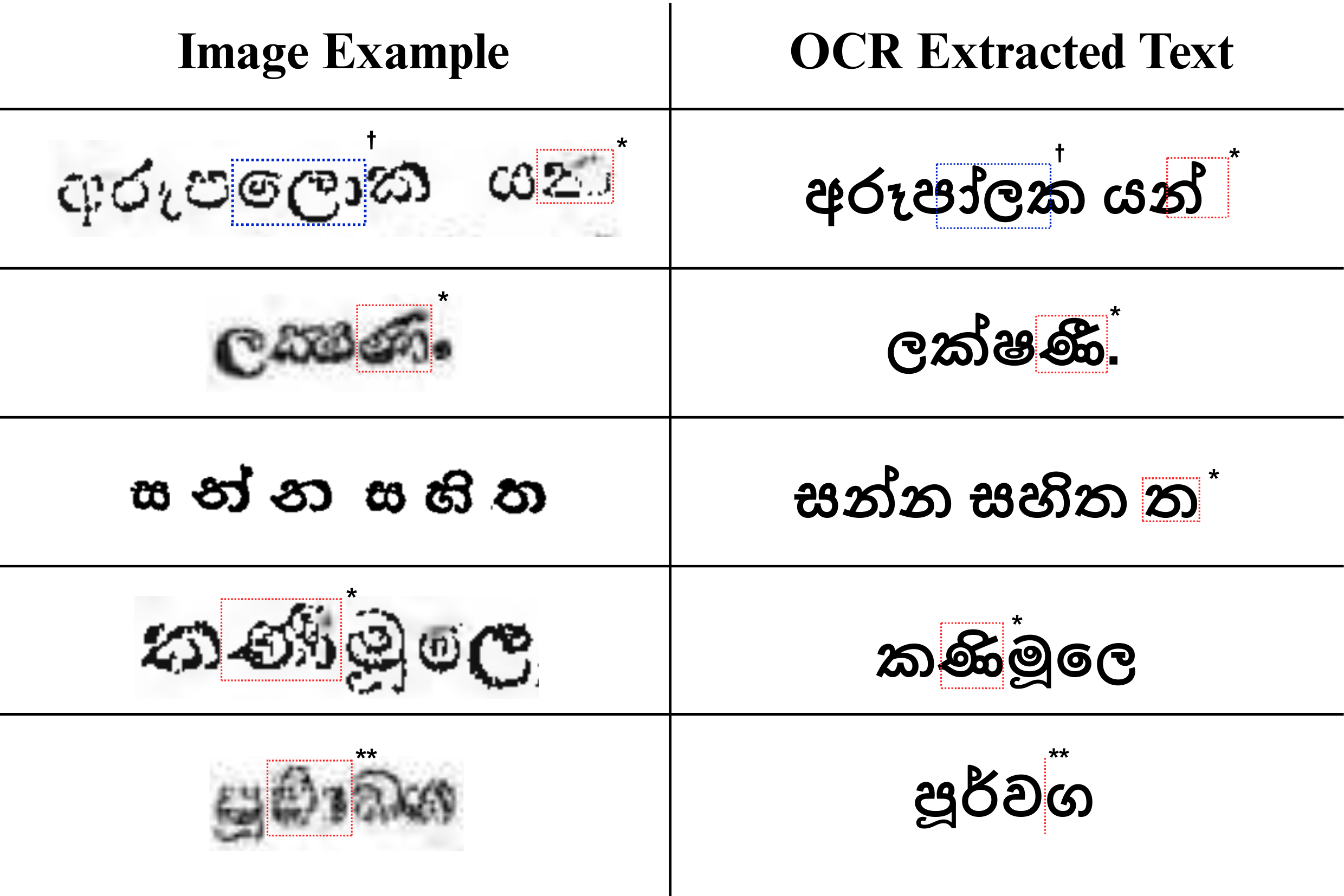}%
    }%
    \caption{Examples of word deformation caused by character-level identification errors, including *incorrect identifications of diacritics/letters and **a complete deletion of a character at the marker. $\dagger$ Note that this is not an error, but rather a text modernisation step.}
    \label{fig:word_and_character_level_errors}
\end{table}

It became clear that character identification issues persisted throughout the documents. As illustrated in Table~\ref{fig:word_and_character_level_errors}, diacritics in Sinhala posed significant challenges. Some characters were completely unrecognised, resulting in character deletions. Additionally, some identified diacritics were incorrect substitutions for different diacritics. There were also instances where a diacritic was erroneously added even when no corresponding character existed, indicative of character insertion errors.
In simpler terms, most of the errors in this category are related to spelling issues. 


\end{document}